\documentclass[11pt]{article}
\usepackage[]{acl}
\usepackage{times}
\usepackage{latexsym}
\usepackage[T1]{fontenc}
\usepackage[utf8]{inputenc}
\usepackage{microtype}
\usepackage{inconsolata}
\usepackage{graphicx}

\usepackage{booktabs}
\usepackage{multirow}
\usepackage{amsmath}
\usepackage{colortbl}
\usepackage{xcolor}
\usepackage{cancel}
\usepackage{paralist}
\usepackage{tabularx}
\usepackage{amssymb}

\usepackage{tikz}
\usepackage[most]{tcolorbox}
\usepackage{dblfloatfix}

\usepackage{float}

\usepackage{enumitem}

\usepackage{tikz}
\usetikzlibrary{shapes.geometric, arrows.meta, positioning, shadows, fit, backgrounds}

\usepackage{graphicx}
\usepackage{longtable} 
\usepackage{array}     
\usepackage{booktabs}  

\usepackage{pifont}

\newcommand{\FN}{\textbf{\textcolor{black}{FN}}}
\newcommand{\FP}{\textbf{\textcolor{black}{FP}}}

\definecolor{bestgreen}{HTML}{E6F4EA} 
\newcommand{\best}[1]{\cellcolor{bestgreen}\textbf{#1}}

\newcommand{\sysQ}{\textsc{Qwen-3}}
\newcommand{\sysPx}{\textsc{Pixtral}}
\newcommand{\sysI}{\textsc{InternVL-3.5}}
\newcommand{\sysGfour}{\textsc{GPT-4.1-mini}}
\newcommand{\sysGfive}{\textsc{GPT-5-mini}}
\newcommand{\sysUCOT}{\textsc{U-CoT+}}
\newcommand{\lsv}[1]{\textbf{\textsc{LSV}}}
\newcommand{\fbhm}[1]{\textbf{FBHM}}
\newcommand{\fhm}[1]{\textsc{FHM}}
\newcommand{\mami}[1]{\textsc{MAMI}}
\newcommand{\fsft}[1]{\textsc{FHM-Sft}}
\newcommand{\msft}[1]{\textsc{MAMI-Sft}}

\newcommand{\ckptQ}{\texttt{Qwen/Qwen3-VL-8B-Instruct}}
\newcommand{\ckptPx}{\texttt{mistral-community/pixtral-12b}}
\newcommand{\ckptI}{\texttt{OpenGVLab/InternVL3\_5-8B}}

\author{
\textbf{Paramananda Bhaskar\textsuperscript{1*}},
\textbf{Naquee Rizwan\textsuperscript{1*}},
\textbf{Daksh Jogchand\textsuperscript{1}},\\
\textbf{Saurabh Kumar Pandey\textsuperscript{2}},
\textbf{Animesh Mukherjee\textsuperscript{1}}\\
{\{pbhaskar, nrizwan, daksh.jogchand\}@kgpian.iitkgp.ac.in}\\
{saurabh2000.iitkgp@gmail.com,\ animeshm@cse.iitkgp.ac.in}\\
\\
\textsuperscript{1}Indian Institute of Technology (IIT), Kharagpur,\quad
\textsuperscript{2}Microsoft\\
{\textbf{\textsuperscript{*}}Equal contribution}
}

\title{\fbhm{}: Functional Benchmarking and Steering of VLMs for Hateful Meme Detection}

\begin{document}
\maketitle

%%%%%%%%%%%%%%%%%%%%%%%%%%%%%%%%%%%%%%%%%%%%%%%

\begin{abstract}
Hateful meme detection remains a formidable challenge for vision-language models, as existing benchmarks are structurally observational---confounding rhetorical hate mechanisms with target community features and preventing causal evaluation of model vulnerabilities. To address this, we introduce \fbhm{}, a systematically curated benchmark of \textit{\textbf{F}unctionality \textbf{B}ased \textbf{H}ateful \textbf{M}emes} constructed along two orthogonal axes: 25 distinct rhetorical functionalities and 10 target communities (5,000 memes total). Benchmarking state-of-the-art VLMs reveals a severe generalization gap: models highly accurate on standard datasets catastrophically drop to near-random performance on \fbhm{}, proving they exploit dataset-specific heuristics rather than robust multimodal reasoning. To efficiently close this gap, we propose \lsv{} (\textit{\textbf{l}earnable \textbf{s}teering \textbf{v}ectors}), an ultra-low data regime strategy that applies a causal intervention objective on as few as 500 steering samples (50 unique base memes), boosting \fbhm{} performance by $\sim$30 Macro-F1 points while outperforming in-context learning and \textsc{PEFT} without degrading source-domain performance. \textit{\textbf{\textcolor{red}{Warning: Contains potentially toxic contents.}}}
\end{abstract}

%%%%%%%%%%%%%%%%%%%%%%%%%%%%%%%%%%%%%%%%%%%%%%%

\section{Introduction}

\begin{figure*}[t]
  \centering
  \includegraphics[width=1\textwidth]{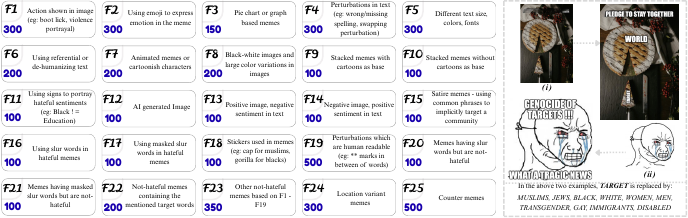}
  \caption{\texttt{Left:} suite of 5,000 FBHM memes spread across 25 functionalities. Each tile presents the functionality number, its description and the corresponding number of memes in that functionality. \texttt{Right:} examples of constructing ten memes for ten target communities using one base image. Three examples for each functionality is provided in Appendix~\ref{sec:appendix_examples}.}
  \label{fig:paradigm}
\end{figure*}

The proliferation of memes on social media has introduced a highly complex vector for the dissemination of hate speech.
Existing standard hate reasoning datasets, such as the \textit{Facebook Hateful Memes} (\fhm{}) \cite{kiela2020hateful} and \mami{} \cite{fersini2022mami} have catalyzed initial research in hate meme detection and control but suffer from \textit{observational entanglement}~\cite{Rizwan_Bhaskar_Das_Majhi_Saha_Mukherjee_2025}. These datasets focus on hate classification without structurally categorizing the underlying semantic or rhetorical mechanisms (examples may include: sentiment mismatches, masked slurs, stacked visual logic, etc.) used to convey the hate. Furthermore, target communities are unevenly distributed in these datasets. Consequently, if a VLM struggles to classify a meme attacking target community $A$ but succeeds on community $B$, the observational nature of the dataset makes it impossible to isolate the root cause. We make our dataset and code public to the research community\footnote{{GitHub: \url{https://github.com/hate-alert/fbhm}\\FBHM dataset: \url{https://huggingface.co/datasets/nrizwan/FBHM}}}.

\noindent In order to bridge these gaps, we propose the \fbhm{} (\textbf{F}unctionality \textbf{B}ased \textbf{H}ateful \textbf{M}emes) dataset (Figure~\ref{fig:paradigm}). Inspired by the principles of functional testing in software engineering~\cite{beizer1996black} and prior works on functionally evaluating LMs in NLP and hate speech detection~\cite{ribeiro-etal-2020-beyond,rottger-etal-2021-hatecheck}, \fbhm{} allows for systematic causal analysis of VLM's behavior by independently varying the target community while keeping the meme's base functionality and image-text structure rigidly constant. By constructing the dataset along two orthogonal axes, 25 carefully chosen functional mechanisms and 10 distinct target communities - \fbhm{} provides a rigorous diagnostic tool to uncover functionality-wise and target-wise disparities in multimodal reasoning.

\noindent VLMs fine-tuned on standard datasets (like \fhm{}, and \mami{}) catastrophically drop to near-random performance on \fbhm{}, while traditional low-resource adaptation strategies--- \textsc{PEFT} with \textsc{LoRA}~\cite{hu2022lora}, and few-shot \textsc{\textsc{ICL}}---all fail to bridge this gap under the extreme low-data constraint of 500 steering samples. To overcome this, we propose \lsv{} (\textbf{l}earnable \textbf{s}teering \textbf{v}ectors), which optimizes layer-wise continuous vectors via a causal intervention objective~\cite{liu2023context,peng2024live} to steer the frozen VLM's representations without weight updates. \lsv{} boosts \fbhm{} performance to $\sim$74--75 Macro-F1, outperforming both \textsc{\textsc{ICL}} and \textsc{PEFT}, while preserving performance on the original \fhm{} and \mami{} benchmarks. In summary, multi-fold contributions of this work are as follows.

\begin{tcolorbox}[title={Key contributions}]
\footnotesize

$\bullet$ We introduce a novel diagnostic benchmark \fbhm{}, a causally-structured dataset of 5,000 memes rigorously curated across 25 rhetorical functionalities and 10 target communities, enabling isolated counterfactual evaluation of VLMs.

$\bullet$ Through extensive benchmarking of open-source and proprietary SOTA models (\sysQ{}, \sysPx{}, \sysI{}, \sysGfour{}, \sysGfive{}, and \sysUCOT{}), we reveal that high in-domain performance on existing datasets (\fhm{}/\mami{}) masks a disastrous failure to generalize to diverse functional structures.

$\bullet$ We demonstrate that standard adaptation techniques like few-shot \textsc{\textsc{ICL}} and \textsc{PEFT} are inherently unsuitable for aligning multimodal reasoning on complex hate memes when restricted to a minimal steering set of 500 samples in scarce data regime.

$\bullet$ We introduce \lsv{}, an activation steering method for efficient alignment by adapting prior in-context vector methods with a causal intervention objective. We show that \lsv{} successfully recovers the generalization gap (achieving up to 30+ Macro-F1 improvement on \fbhm{}) uniformly across all target communities and functionalities.

\end{tcolorbox}

%%%%%%%%%%%%%%%%%%%%%%%%%%%%%%%%%%%%%%%%%%%%%%%

\section{Related work}

\textbf{Hateful memes}: Early efforts to combat multimodal hate has yielded in datasets like \textit{Facebook Hateful Memes} (\fhm{})~\cite{kiela2020hateful},  \mami{} (misogynistic memes) \cite{fersini2022mami}, \textsc{HARM} (harmful memes on COVID-19 \& \textsc{US-politics})~\cite{pramanick2021detecting}, \textsc{GOAT-Bench}~\cite{lin2024goat}, and \textsc{ToxicTAGS}~\cite{stemtox} among others.
Some works provide explanation and intervention upon hateful memes~\cite{Hee_Lee_2025,jha-etal-2024-memeguard,adak2025memesense,rizwan2026see}.
Works such as \textsc{HaTReD}~\cite{hee2023hatred} also have target annotations.\\
\noindent\textbf{Functional evaluation of language models}:
In the text domain, prior works have systematically stress-tested LLMs; for instance, \citet{das2023evaluatingchatgptsperformancemultilingual} evaluated ChatGPT's zero-shot robustness across complex multilingual and emoji-based hate speech to understand how models process implicit textual toxicity. \cite{ribeiro-etal-2020-beyond} proposed \textsc{CheckList} for behavioral testing of NLP models. \cite{zhao2022vl} proposed dataset similar to \textsc{CheckList} for vision models. \cite{thrush2022winoground} also proposed an evaluation benchmark for multimodal reasoning in VLMs. Despite these theoretical advancements, standardizing functional evaluation for VLMs still remains an open challenge.\\
\noindent\textbf{Alignment of VLMs}:
Supervised fine-tuning (\textsc{Sft}) of VLMs is expensive and is prone to catastrophic forgetting~\cite{10.5555/3600270.3602446} and low-rank adaptation (\textsc{LoRA})~\cite{hu2021lora} don't deliver promising results in low-data regime. The concept of in-context vectors (\textsc{ICV}s) was introduced in~\cite{liu2023context} as a training-free method to steer language models by adding task-specific vectors to hidden states. \cite{peng2024live} extended this to VLMs with \textit{learnable in-context vectors}, which learn continuous vectors using only a small set of demonstrations.

%%%%%%%%%%%%%%%%%%%%%%%%%%%%%%%%%%%%%%%%%%%%%%%

\section{\fbhm{} dataset}
\label{sec:annotation_fbhm}
The mutlimodal \fbhm{} dataset is constructed along two orthogonal axes to ensure unbiased and controlled evaluations: \textit{functionalities} and \textit{target communities}. \fbhm{} dataset features 25 carefully curated functionalities (inspired by \textsc{HateCheck}~\citet{rottger-etal-2021-hatecheck}) and 10 target communities.

\noindent \textbf{Functionalities}: The functionalities were designed collaboratively by four experienced researchers actively working in this domain. Anchored in prior works on textual hate speech~\cite{rottger-etal-2021-hatecheck, das2023evaluatingchatgptsperformancemultilingual}, these functionalities target specific multimodal structures consistently mishandled by current models.
As summarized in Figure~\ref{fig:paradigm}, the 25 functionalities span five conceptually distinct dimensions as follows - (a) visual formats and imagery types (F1, F3, F7, F8, F12), (b) textual obfuscation and lexical evasion (F4, F5, F16, F17, F19), (c) structural composition and visual metaphor (F2, F9, F10, F11, F18), (d) pragmatic inference and sentiment misalignment (F6, F13, F14, F15, F24), and (e) not-hateful contrast and counterspeech (F20, F21, F22, F23, F25).

\noindent \textbf{Target communities}: To ensure that \fbhm{} serves as a robust benchmark for generalization, we utilize 10 target communities to scale the dataset across these 10 distinct groups uniformly. The protected groups considered in this study are \textit{Muslims, Jews, Black, White, women, men, transgender, gay, immigrants,} and \textit{disabled/down syndrome} individuals. This uniform representation prevents the model from exploiting demographic-specific lexical biases and ensures that our 25 functionalities are evaluated against a diverse spectrum of cultural and social identities.

\noindent\textbf{Dataset curation}: The core design principle for creating \fbhm{} is to create a base image-text pair demonstrating a specific \textit{functionality} (see Figure~\ref{fig:paradigm}) and then change \textit{only} the target community in the image to generate 10 controlled variants per base meme. This structure facilitates causal-style comparisons: evaluating across targets given the same functionality, and evaluating across functionalities given the same target community. Refer Appendices~\ref{sec:image_sources} and \ref{sec:appendix_examples} for more details.

\noindent \textbf{Template selection}: The template design underwent a rigorous process before annotation. Prior to dataset curation, annotators spent approximately one month familiarizing themselves with contemporary meme trends across multiple social media platforms. We subsequently prepared a comprehensive 40-page guideline documenting diverse meme mechanisms, rhetorical structures, and target interactions. This process identified around 800 candidate base meme template types. The four in-domain researchers jointly reviewed these candidates and unanimously selected 500 base templates that were judged to generalize appropriately across their intended target categories.

\noindent\textbf{Annotation process}: All memes were created and annotated by an experienced team of four domain-expert researchers, with each instance binary-labeled as hateful \texttt{(1)} or not-hateful \texttt{(0)} following a detailed rubric covering both explicit indicators (slurs, dehumanizing language) and implicit hate (sarcasm, stereotypes, mocking). To validate label quality, all 500 base memes were independently classified by two experts, yielding a Cohen's $\kappa$ of 0.84 and confirming highly reliable annotations.

\noindent\textbf{Dataset statistics and splits:} The final \fbhm{} dataset consists of 5,000 memes, derived from 500 base images, each replicated across the 10 target communities. To create a \textit{steering set} ($\mathcal{D}_{\text{steer}}$) for parameter-efficient intervention, we utilize stratified random sampling ($\sim$10\%, 50 base images $\times$ 10 targets = 500 samples). We attribute the remaining 4,500 samples as the \textit{test set} for benchmarking, strictly preserving the base-variant integrity. The label distributions are detailed in Table \ref{tab:dataset_stats}. The combined dataset features a deliberate 67\% hateful and 33\% not-hateful split, reflecting the realistic skew of functionally targeted edge cases in moderation queues—where automated pre-filtering heavily concentrates likely violations for human review \citep{enforcement-twitter-2026}—and directly aligns with the $\sim$68.8\% hateful to 31.2\% not-hateful distribution established in foundational functional test suites like \textsc{HateCheck} \citep{rottger-etal-2021-hatecheck}.

\begin{table}[t]
\centering
\footnotesize
\begin{tabular}{@{}llcc@{}}
\toprule
\textbf{split} & \textbf{label} & \textbf{count} & \textbf{percentage} \\ \midrule
\multirow{2}{*}{steering set} & not-hateful & 150 & 30.0\% \\
 & hateful & 350 & 70.0\% \\ \midrule
\multirow{2}{*}{test set} & not-hateful & 1,500 & 33.3\% \\
 & hateful & 3,000 & 66.7\% \\ \midrule
\multirow{2}{*}{\textbf{combined}} & \textbf{not-hateful} & \textbf{1,650} & \textbf{33.0\%} \\
 & \textbf{hateful} & \textbf{3,350} & \textbf{67.0\%} \\ \bottomrule
\end{tabular}
\caption{\fbhm{} dataset split and label distribution.}
\label{tab:dataset_stats}
\end{table}

%%%%%%%%%%%%%%%%%%%%%%%%%%%%%%%%%%%%%%%%%%%%%%%
\setlength{\belowdisplayskip}{1pt} \setlength{\belowdisplayshortskip}{1pt}
\setlength{\abovedisplayskip}{1pt} \setlength{\abovedisplayshortskip}{1pt}
\section{\lsv{} formulation}
\label{sec:lsv_formulation}
Building upon the mathematical framework of activation steering \cite{peng2024live}, we introduce a parameter-efficient intervention mechanism on a frozen target VLM, denoted as $\mathcal{M}$. Let $L$ represent the total number of transformer layers in $\mathcal{M}$. We initialize a learnable intervention space consisting of a set of layer-wise steering vectors $V = \{\mathbf{v}_1, \mathbf{v}_2, \dots, \mathbf{v}_L\}$, where $\mathbf{v}_l \in \mathbb{R}^d$ matches the hidden dimension of the model. To regulate the layer-specific magnitude of these perturbations, we jointly learn a set of scalar coefficients $\alpha = \{\alpha_1, \alpha_2, \dots, \alpha_L\}$. For a given input query $x_q$, the modified hidden state $\mathbf{h}_l'$ at layer $l$ is formulated as:
\begin{equation}\label{eq:1}
    \small
    \mathbf{h}_l'(x_q) = \mathbf{h}_l(x_q) + c \cdot \alpha_l \cdot \mathbf{v}_l, \quad \forall\; l \in \{1,\dots,L\}
\end{equation}
where $\mathbf{h}_l$ is the original, unperturbed hidden state. The scaling constant $c \in \mathbb{R}_{>0}$ serves as a tunable inference-time scaling hyperparameter. During the training phase, $c$ is strictly set to $1$. The parameters $V$ and $\alpha$ are optimized exclusively on our minimal steering set $\mathcal{D}_{\text{steer}}$, containing $N=500$ paired samples $(x_q, y)$ (originating from just 50 unique base images). To ensure the steered model robustly maps to the desired functional behaviors, we design a dual-objective loss function. First, we construct a reference distribution. For each query $x_q$, we randomly sample $k=32$ exemplars from $\mathcal{D}_{\text{steer}} - \{x_q\}$ to construct an \textsc{ICL} demonstration prompt, $X_{\text{demo}}$. We forward $X_{\text{demo}}$ through the unmodified base model $\mathcal{M}$ to extract the reference probability distribution over the entire vocabulary $\mathcal{V}$ for the \textit{first generated token}. We denote this vocabulary-wide distribution as $P_{\text{demo}}(\mathcal{V}) = \mathcal{M}(\mathcal{V} \mid X_{\text{demo}})$. Notably, we employ a unified, class-agnostic sampling strategy where 32 exemplars are drawn at random from $\mathcal{D}_{\text{steer}} - \{x_q\}$. This stochastic mixture of hateful and not-hateful memes is presented in a single, random order within the prompt $X_{\text{demo}}$. Concurrently, we forward the isolated query $x_q$ through the intervened model (denoted as $\mathcal{M}_{V, \alpha}$) to obtain the steered vocabulary distribution over the entire vocabulary $\mathcal{V}$ for the \textit{first generated token}, denoted as $P_{\text{lsv}}(\mathcal{V}) = \mathcal{M}_{V, \alpha}(\mathcal{V} \mid x_q)$. To transfer the cognitive reasoning of the \textsc{ICL} prompt into the vectors, we minimize the \textsc{Kullback-Leibler} (KL) divergence between these distributions:
\begin{equation}\label{eq:2}
    \small
    \mathcal{L}_d = \mathbb{E}_{x_q \sim \mathcal{D}_{\text{steer}}} \left[ D_{\text{KL}}\big( P_{\text{demo}}(\mathcal{V}) \;\|\; P_{\text{lsv}}(\mathcal{V}) \big) \right]
\end{equation}
Second, to strictly anchor the semantic trajectory of the steering vectors to the correct downstream classification task, we apply a ground-truth intervention loss via cross-entropy:
\begin{equation}\label{eq:3}
    \small
    \mathcal{L}_{\text{gt}} = \mathbb{E}_{(x_q, y) \sim \mathcal{D}_{\text{steer}}} \left[ -\log \Big( P_{\text{lsv}}(\mathcal{V})\big[\text{token}(y)\big] \Big) \right]
\end{equation}
where $\text{token}(y)$ is a deterministic function that maps the ground-truth string label $y$ to its corresponding initial vocabulary token index, allowing us to query the exact probability scalar from the distribution $P_{\text{lsv}}(\mathcal{V})$. The final optimization objective is a combination of the distribution alignment and the ground-truth anchor as follows.
\begin{equation}\label{eq:4}
    \small
    \mathcal{L}_{\text{total}} = \mathcal{L}_d + \lambda \cdot \mathcal{L}_{\text{gt}}
\end{equation}
where $\lambda$ is a hyperparameter regulating the relative importance of the classification loss. Following the foundational framework of \citeauthor{peng2024live} \shortcite{peng2024live}, we empirically set $\lambda = 0.5$. Further, unlike prior approaches that compute loss expectations across the entire auto-regressive generation sequence, our formulation strictly constrains $\mathcal{L}_{\text{total}}$ to the \textit{first decoded token}. For sequence classification tasks framed as auto-regressive generation (for eg: predicting `hateful' vs `not-hateful'), the sequence-averaged approach violates the strict causal structure of the underlying model. By localizing our objective, we establish a rigorous causal intervention mechanism, ensuring the steering vectors unequivocally dictate the model's initial functional classification logic without suffering from auto-regressive variance, sequence dilution, or tokenizer-specific artifacts (such as the multi-token fragmentation of the word `not-hateful'). We provide further details in Appendix~\ref{sec:ablation_first_token}.

\noindent\textbf{Inference-time scaling.} At inference, the base model $\mathcal{M}$ remains entirely frozen, and no \textsc{\textsc{ICL}} demonstrations are provided. We inject the learned parameters $V$ and $\alpha$ directly into the zero-shot forward pass using Equation~\ref{eq:1}. To maximize the robustness of the latent shift, the multiplier $c$ serves as a tunable inference hyperparameter. For our evaluations, $c$ is swept across the interval $[0.5, 2.5]$ in increments of 0.1 (yielding 21 discrete values). Because the optimal $c$ is highly architecture- and domain- dependent, it is selected strictly using small, domain-specific held-out validation splits—$\mathcal{D}_{\text{steer}}$ (500 samples) for \fbhm{}, the \fhm{} \texttt{dev\_seen} split (500 samples), and the \mami{} validation set (1,000 samples)—requiring zero access to the test distribution. We analyze this scaling behavior in detail in Section~\ref{sec:results}.

%%%%%%%%%%%%%%%%%%%%%%%%%%%%%%%%%%%%%%%%%%%%%%%

\section{Experimental setup}

\begin{table}[t]
\centering
\footnotesize
\setlength{\tabcolsep}{4pt}
\begin{tabular}{@{}ll r rr@{}}
\toprule
\textbf{dataset} & \textbf{split} & \textbf{total} & \textbf{class 0} & \textbf{class 1} \\ 
\midrule

\multirow{3}{*}{\textbf{FHM}} 
 & train & 8,500 & 5,481 & 3,019 \\
 & dev   & 500   & 253   & 247 \\
 & test  & 1,000 & 510   & 490 \\ 

\midrule

\multirow{3}{*}{\textbf{MAMI}} 
 & train & 9,000 & 4,500 & 4,500 \\
 & val   & 1,000 & 500   & 500 \\
 & test  & 1,000 & 500   & 500 \\ 

\bottomrule
\end{tabular}
\caption{Label distribution of standard benchmark datasets (\fhm{} and \mami{}). Class 0 denotes benign samples; class 1 denotes hateful/misogynistic samples.}
\label{tab:external_dataset_stats}
\end{table}

To justify the necessity of \fbhm{}, we demonstrate the limitations of current alignment paradigms and establish the efficacy of \lsv{} technique. We conduct extensive evaluations across a suite of VLMs in zero-shot, few-shot, fine-tuned, and low-data steered settings.

\noindent\textbf{Additional datasets}: In addition to our proposed \fbhm{} dataset, we utilize the standard \fhm{} and \mami{} benchmarks for supervised fine-tuning and evaluation. Table~\ref{tab:external_dataset_stats} summarizes their label distributions. \fhm{} exhibits a class imbalance skewed toward not-hateful samples, while \mami{} is balanced across both training and test splits. However, both datasets are observational and do not provide controlled variation across functionalities and target communities, motivating the need for \fbhm{}.

\noindent\textbf{Employed models}: We evaluate state-of-the-art open-source VLMs, including \sysQ{}~\cite{qwen3} (\ckptQ{}), 
\sysPx{}~\cite{pixtral} (\ckptPx{}), and 
\sysI{}~\cite{internvl} (\ckptI{}), 
alongside closed-source frontier models (\sysGfour{}, \sysGfive{}) and specialized architectures like \sysUCOT{}~\cite{ucotplus}. For each open-source VLM, we define three source-domain baselines as follows.\\ 
\noindent $\bullet$~\textsc{Base}: The pre-trained instruct model evaluated without any task-specific fine-tuning.\\
\noindent $\bullet$~\fsft{}: The model fully fine-tuned on the \fhm{} training set.\\
\noindent $\bullet$~\msft{}: The model fully fine-tuned on the \mami{} training set.

In both the above cases, \textsc{SFT} was performed for 15 epochs with early stopping based on validation loss. All evaluations report accuracy and Macro-F1 (from now on we will call it MF1) scores. Detailed task framing and experimental setup are provided in Appendices~\ref{app:task_framing} and \ref{sec:implementation_details} respectively.

%%%%%%%%%%%%%%%%%%%%%%%%%%%%%%%%%%%%%%%%%%%%%%%

\section{Computational efficiency}
\label{sec:compute}

We compare the four paradigms--\textsc{zero-shot}, \textsc{ICL}, \textsc{PEFT} (\textsc{QLoRA}), and \lsv{} along three axes: trainable parameters, inference-time token count, and GPU memory footprint.

\noindent\textbf{(A) Trainable parameters} -- \textbf{(i) \textsc{zero-shot} / \textsc{ICL}}: Parameter tuning-free; no weight updates occur. \textbf{(ii) \textsc{PEFT}}: \textsc{QLoRA} adapters are injected into 7 projection modules per layer. For rank $r{=}16$, hidden dim $d{=}4,096$, and $L{=}32$ layers:
\begin{equation}
\small
N_{\text{PEFT}} = 7 \times d \times 2r \times L
               \approx 29.4\text{M}
\end{equation}

\noindent\textbf{(iii) \lsv{}}: One steering vector $\mathbf{v}_l \in \mathbb{R}^d$ and scalar $\alpha_l$ per layer:
\begin{equation}
\small
N_{\text{LSV}} = L(d+1) \approx 131\text{K}
\end{equation}
As a conclusion, \lsv{} requires ${\approx} 224$ times fewer parameters than \textsc{PEFT}. Table~\ref{tab:param_count} summarizes the trainable parameters.

\begin{table}[t]
\centering
\scriptsize
\renewcommand{\arraystretch}{1.25}
\resizebox{\columnwidth}{!}{
\begin{tabular}{@{}lrcc@{}}
\toprule
\textbf{paradigm} & \textbf{params} & \textbf{wt.\ update} & \textbf{backbone} \\
\midrule
\textsc{zero-shot}   & 0        & \ding{55} & frozen \\
\textsc{ICL}    & 0        & \ding{55} & frozen \\
\textsc{PEFT}   & 29.4M    & \ding{51} & partial \\
\lsv{}          & 131K     & \ding{55} & frozen \\
\bottomrule
\end{tabular}
}
\caption{Trainable parameters count. Here, $L{=}32$, $d{=}4096$, $r{=}16$. Refer Section~\ref{sec:compute} (A) for further details.}
\label{tab:param_count}
\end{table}

\noindent\textbf{(B) Inference-time token count} -- Token count governs KV-cache size and attention compute. Using our 
prompt template $\mathcal{T}$ (refer to Appendix~\ref{app:task_framing}): system 
prompt $|S|{\approx}120$ tokens (shared once) and query $|U(x)|{\approx}60$ tokens.

\noindent\textbf{(i) \textsc{Base}/\textsc{PEFT}/\lsv{}} operate zero-shot:
\begin{equation}
\small
T_{\text{zs}} = |S|+|U(x)| \approx 180\ \text{tokens}
\end{equation}

\noindent\textbf{(ii) \textsc{ICL}} prepends $k{=}32$ demonstrations. Each 
demonstration contributes user query $|U(x_i)|{\approx}60$ tokens and assistant 
response $|A(y_i)|{\approx}5$ tokens; $S$ is shared once:
\begin{equation}
\small
\begin{split}
T_{\text{ICL}} &= |S| + k\bigl(|U(x_i)|+|A(y_i)|\bigr) + |U(x)| \approx 2{,}260\ \text{tok.}
\end{split}
\end{equation}
At $336{\times}336$ resolution with patch size 14, each image contributes 
${\approx}576$ patch tokens. With 32 demonstration images plus 1 query image, 
visual tokens total $33{\times}576{=}19{,}008$, giving an aggregate of 
$2{,}260{+}19{,}008{=}21{,}268$ tokens per query which is a \textbf{${\approx}28$ times} 
overhead over zero-shot. Table~\ref{tab:token_counts} summarizes these counts.

\begin{table}[t]
\centering
\scriptsize
\renewcommand{\arraystretch}{1.25}
\resizebox{\columnwidth}{!}{
\begin{tabular}{@{}lrrr@{}}
\toprule
\textbf{paradigm} & \textbf{text tok.} & \textbf{visual tok.} & \textbf{total} \\
\midrule
\textsc{zero-shot}   & 180   & 576    & 756    \\
\textsc{PEFT}   & 180   & 576    & 756    \\
\textsc{ICL}    & 2,260 & 19,008 & 21,268 \\
\lsv{}          & 180   & 576    & 756    \\
\bottomrule
\end{tabular}
}
\caption{Inference-time token counts. Here, $k{=}32$; $336{\times}336$ 
images; patch size = 14. Refer Section~\ref{sec:compute} (B) for further details.}
\label{tab:token_counts}
\end{table}

\noindent\textbf{(C) GPU memory at inference} -- We decompose memory into backbone $M_{\text{base}}$, paradigm parameters $M_{\text{param}}$, and KV-cache $M_{\text{kv}}$. An 8B model in 4-bit NF4 precision occupies:
\begin{equation}
\small
M_{\text{base}} = 8\text{B} \times 0.5\ \text{bytes/param} = 4\ \text{GB}
\end{equation}
With activations and runtime buffers, the working baseline is ${\sim}6$--$7$ GB. \textbf{KV-cache} for $L{=}32$ layers, $n_h{=}32$ heads, $d_h{=}128$, \texttt{bfloat16}:
\begin{equation}
\small
\begin{split}
M_{\text{kv}} &= 2 \cdot L \cdot T \cdot n_h \cdot d_h \cdot 2\ \text{B} \approx 0.524T\ \text{MB}
\end{split}
\end{equation}
For \textbf{\textsc{zero-shot}} ($T{\approx}756$): $M_{\text{kv}}{\approx}0.40$ GB. 
For \textbf{\textsc{ICL}} ($T{\approx}21{,}268$): $M_{\text{kv}}{\approx}11.1$ GB, pushing 
total inference memory to ${\sim}17$ GB at batch size 1 and risking OOM at batch size ${>}4$ on 40 GB GPUs. For \textbf{\textsc{PEFT}}, adapter weights in \texttt{bfloat16} add $29.4\text{M}{\times}2{\approx}58.8$ MB. Finally for \textbf{\lsv{}}, vectors and scalars add $131\text{K}{\times}2{\approx}0.26$ MB (${\sim}226$ times less than \textsc{PEFT}). Refer Table~\ref{tab:gpu_memory} for details.

\begin{table}[t]
\centering
\scriptsize
\renewcommand{\arraystretch}{1.25}
\resizebox{\columnwidth}{!}{
\begin{tabular}{@{}lrrrc@{}}
\toprule
\textbf{paradigm} & \textbf{$M_{\text{param}}$} & \textbf{$M_{\text{kv}}$} & \textbf{total} & \textbf{OOM risk} \\
\midrule
\textsc{zero-shot}   & 0 MB      & 0.40 GB & 7.0 GB  & low  \\
\textsc{ICL}    & 0 MB  & 11.1 GB & 17.4 GB & mod--high \\
\textsc{PEFT}   & 58.8 MB   & 0.40 GB & 7.5 GB  & low  \\
\lsv{}          & 0.26 MB   & 0.40 GB & 7.1 GB  & low  \\
\bottomrule
\end{tabular}
}
\caption{GPU memory at inference time. Here, $M_{\text{base}}{\approx}6$--$7$ GB; batch size = 1; \texttt{bfloat16} KV-cache; 40 GB GPU. Refer Section~\ref{sec:compute} (C) for further details.}
\label{tab:gpu_memory}
\end{table}

\begin{table}[t]
\centering
\scriptsize
\renewcommand{\arraystretch}{1.25}
\resizebox{\columnwidth}{!}{
\begin{tabular}{@{}lrrrr@{}}
\toprule
\textbf{paradigm} & \textbf{params} & \textbf{tokens} & \textbf{extra mem.} & \textbf{MF1} \\
\midrule
\textsc{zero-shot}   & 0       & 756    & 0 MB     & 46--47 \\
\textsc{ICL}    & 0  & 21,268 & 11.1 GB & 48--50 \\
\textsc{PEFT}   & 29.4M   & 756    & 58.8 MB  & 46--47 \\
\lsv{}          & 131K    & 756    & 0.26 MB  & \textbf{72--75} \\
\bottomrule
\end{tabular}
}
\caption{Unified comparison: \lsv{} achieves the highest \fbhm{} MF1 with zero-shot token overhead, $224$ times fewer parameters than \textsc{PEFT}, and $>50{,}000$ times less KV-cache than \textsc{ICL}. Refer Section~\ref{sec:compute} (D) for further details.}
\label{tab:compute_summary}
\end{table}

\noindent\textbf{(D) Conclusive benefits} -- \lsv{} occupies a uniquely favorable position across all three axes. \textsc{ICL}, though parameter-free, incurs prohibitive memory costs from its $33\times$ token expansion. \textsc{PEFT} remains inference-efficient but fails statistically: 500 samples are insufficient to generalize adapter weights across 25 functionalities. \lsv{} bypasses both bottlenecks by distilling reasoning into ${\approx}131$K continuous vectors, achieving up to $+30$ MF1 on \fbhm{} at negligible computational overhead. Concrete details are in Table~\ref{tab:compute_summary}.

%%%%%%%%%%%%%%%%%%%%%%%%%%%%%%%%%%%%%%%%%%%%%%%%%%%%%%%%%%%%

\section{Results and analysis}
\label{sec:results}

\begin{table*}[!t]
\centering
\footnotesize
\renewcommand{\arraystretch}{1}
\setlength{\tabcolsep}{3mm}
\begin{tabular}{@{}ll l cc ||cc cc@{}}
\toprule
\multirow{2}{*}{\textbf{model}} & \multirow{2}{*}{\textbf{variant}} & \multirow{2}{*}{\textbf{setup}} & \multicolumn{2}{c}{\textbf{\fbhm{} test}} & \multicolumn{2}{c}{\textbf{\fhm{} test}} & \multicolumn{2}{c}{\textbf{\mami{} test}} \\
\cmidrule(lr){4-5} \cmidrule(lr){6-7} \cmidrule(lr){8-9}
 & & & acc & mf1 & acc & mf1 & acc & mf1 \\
\midrule

% --- QWEN 3 ---
\multirow{12}{*}{\sysQ{}} 
 & \multirow{4}{*}{\textsc{Base}} 
 & \textsc{Base} & 54.78 & 45.99 & 68.60 & 68.33 & 78.70 & 78.64 \\
 & & + \textsc{PEFT} & 54.22 & 46.64 & 68.30 & 68.00 & \best{79.10} & \best{79.10} \\
 & & + \textsc{\textsc{ICL}} & 56.22 & 48.65 & 66.80 & 65.84 & 79.00 & 79.00 \\
 & & \textbf{\textcolor{blue!80!black}{+ \lsv{}}} & \best{76.09} & \best{72.78} & \best{69.80} & \best{69.79} & 77.60 & 77.49 \\
\cmidrule(l){2-9}

 & \multirow{4}{*}{\fsft{}} 
 & \textsc{Base} & 54.80 & 45.53 & 74.80 & 74.05 & 74.10 & 74.10 \\
 & & + \textsc{PEFT} & 54.47 & 46.34 & 68.90 & 68.64 & \best{79.60} & \best{79.59} \\
 & & + \textsc{\textsc{ICL}} & 53.22 & 49.11 & 70.70 & 69.14 & 75.60 & 75.57 \\
 & & \textbf{\textcolor{blue!80!black}{+ \lsv{}}} & \best{77.82} & \best{74.46} & \best{76.20} & \best{75.98} & 72.30 & 72.29 \\
\cmidrule(l){2-9}

 & \multirow{4}{*}{\msft{}} 
 & \textsc{Base} & 55.27 & 46.78 & 70.40 & 70.17 & 80.40 & 80.19 \\
 & & + \textsc{PEFT} & 54.78 & 46.73 & 68.70 & 68.46 & 80.20 & 80.20 \\
 & & + \textsc{\textsc{ICL}} & 56.58 & 49.64 & 67.90 & 66.71 & \best{81.90} & \best{81.80} \\
 & & \textbf{\textcolor{blue!80!black}{+ \lsv{}}} & \best{78.42} & \best{75.44} & \best{72.80} & \best{72.80} & 79.40 & 78.89 \\
\midrule

% --- PIXTRAL ---
\multirow{12}{*}{\sysPx{}} 
 & \multirow{4}{*}{\textsc{Base}} 
 & \textsc{Base} & 65.38 & 41.53 & 61.40 & 58.74 & 67.42 & 63.79 \\
 & & + \textsc{PEFT} & \best{66.27} & 41.54 & 58.40 & 53.18 & 65.14 & 60.57 \\
 & & + \textsc{\textsc{ICL}} & 61.00 & 43.57 & 68.40 & 68.09 & 76.67 & 76.39 \\
 & & \textbf{\textcolor{blue!80!black}{+ \lsv{}}} & 63.21 & \best{58.08} & \best{68.60} & \best{68.55} & \best{80.70} & \best{80.70} \\
\cmidrule(l){2-9}

 & \multirow{4}{*}{\fsft{}} 
 & \textsc{Base} & 65.84 & 41.38 & 59.90 & 54.34 & 72.20 & 71.16 \\
 & & + \textsc{PEFT} & 66.24 & 41.25 & 58.10 & 52.59 & 64.23 & 59.25 \\
 & & + \textsc{\textsc{ICL}} & 61.75 & 46.35 & \best{73.80} & \best{73.35} & 72.91 & 72.49 \\
 & & \textbf{\textcolor{blue!80!black}{+ \lsv{}}} & \best{68.36} & \best{58.79} & 72.40 & 71.54 & \best{75.30} & \best{75.16} \\
\cmidrule(l){2-9}

 & \multirow{4}{*}{\msft{}} 
 & \textsc{Base} & 66.31 & 42.77 & 59.20 & 53.87 & 74.00 & 72.87 \\
 & & + \textsc{PEFT} & 66.27 & 41.32 & 57.70 & 51.91 & 64.06 & 59.04 \\
 & & + \textsc{\textsc{ICL}} & 62.10 & 47.95 & \best{70.20} & 69.75 & 76.00 & 75.19 \\
 & & \textbf{\textcolor{blue!80!black}{+ \lsv{}}} & \best{76.18} & \best{71.59} & \best{70.20} & \best{70.17} & \best{78.50} & \best{78.14} \\
\midrule

% --- INTERNVL ---
\multirow{12}{*}{\sysI{}} 
 & \multirow{4}{*}{\textsc{Base}} 
 & \textsc{Base} & 56.69 & 46.84 & 65.30 & 64.88 & 74.20 & 74.15 \\
 & & + \textsc{PEFT} & 56.31 & 44.95 & 65.60 & 65.33 & \best{74.60} & \best{74.47} \\
 & & + \textsc{\textsc{ICL}} & 55.98 & 48.62 & 63.70 & 62.11 & 71.80 & 71.45 \\
 & & \textbf{\textcolor{blue!80!black}{+ \lsv{}}} & \best{70.04} & \best{52.48} & \best{67.30} & \best{67.29} & 69.70 & 67.83 \\
\cmidrule(l){2-9}

 & \multirow{4}{*}{\fsft{}} 
 & \textsc{Base} & 57.29 & 45.37 & \best{73.20} & \best{73.08} & 72.60 & 71.88 \\
 & & + \textsc{PEFT} & 56.47 & 45.14 & 65.40 & 65.10 & \best{74.60} & \best{74.46} \\
 & & + \textsc{\textsc{ICL}} & 57.80 & 49.79 & 70.50 & 69.81 & 72.40 & 72.16 \\
 & & \textbf{\textcolor{blue!80!black}{+ \lsv{}}} & \best{69.80} & \best{56.55} & 72.80 & 72.57 & 74.30 & 74.04 \\
\cmidrule(l){2-9}

 & \multirow{4}{*}{\msft{}} 
 & \textsc{Base} & 50.00 & 46.06 & 65.20 & 63.59 & \best{78.10} & \best{77.69} \\
 & & + \textsc{PEFT} & 56.31 & 44.94 & 65.30 & 65.00 & 74.50 & 74.35 \\
 & & + \textsc{\textsc{ICL}} & 56.02 & 50.58 & 64.50 & 62.94 & 76.00 & 75.98 \\
 & & \textbf{\textcolor{blue!80!black}{+ \lsv{}}} & \best{67.89} & \best{51.89} & \best{69.60} & \best{69.58} & \best{78.10} & 77.68 \\
\midrule
\midrule

% --- OTHER BASELINES ---
\multicolumn{9}{l}{\textbf{Proprietary and other baselines}} \\
\midrule
\multirow{2}{*}{\sysGfour{}} 
 & zero-shot & \textsc{Base} & 56.62 & 52.39 & 70.20 & 69.97 & \best{81.90} & \best{81.80} \\
 & in-context & + \textsc{\textsc{ICL}} & \best{63.24} & \best{55.17} & \best{72.50} & \best{72.47} & 78.30 & 77.98 \\
\cmidrule(l){2-9}
\multirow{2}{*}{\sysGfive{}} 
 & zero-shot & \textsc{Base} & 58.38 & 50.23 & \best{74.15} & \best{73.62} & \best{84.86} & \best{84.86} \\
 & in-context & + \textsc{\textsc{ICL}} & \best{60.64} & \best{52.73} & 72.80 & 72.17 & 82.20 & 82.14 \\
\cmidrule(l){2-9}
\textsc{U-CoT+} & pre-trained & prompting & 57.76 & 48.41 & 73.40 & 73.39 & 79.90 & 79.89 \\
\bottomrule
\end{tabular}
\caption{Comparison of the different approaches. The proposed \lsv{} approach achieves massive improvements on \fbhm{} while preserving or enhancing source-domain performance. Here, FHM and MAMI test are treated as source-domain test suites. acc: accuracy; mf1: Macro-F1 score.}
\label{tab:alignment_comparison}
\end{table*}

\begin{figure}
  \centering
  \includegraphics[width=0.48\textwidth]{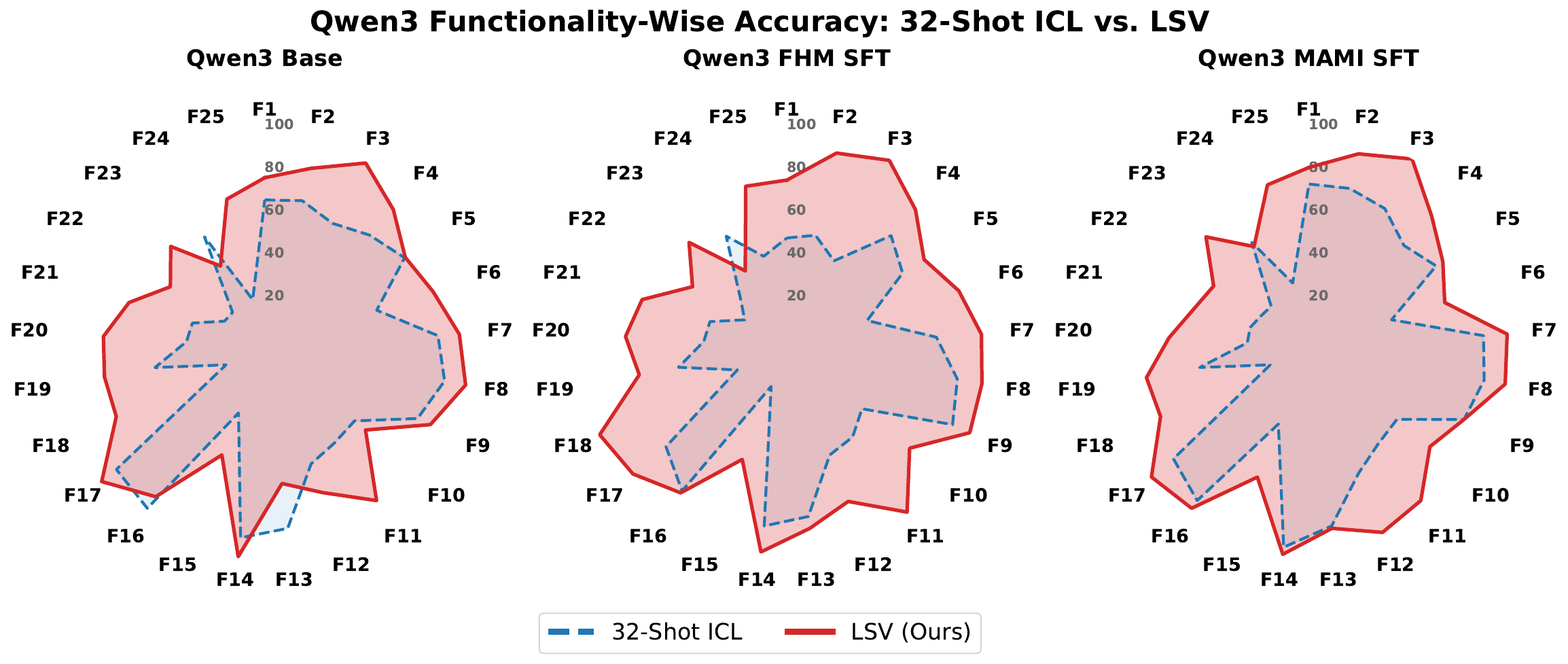}
  \caption{Functionality-wise accuracy comparison between 32-shot \textsc{\textsc{ICL}} and our proposed \lsv{} across three \sysQ{} variants.}
  \label{fig:func_radar}
  \vspace{1em}
  \includegraphics[width=0.48\textwidth]{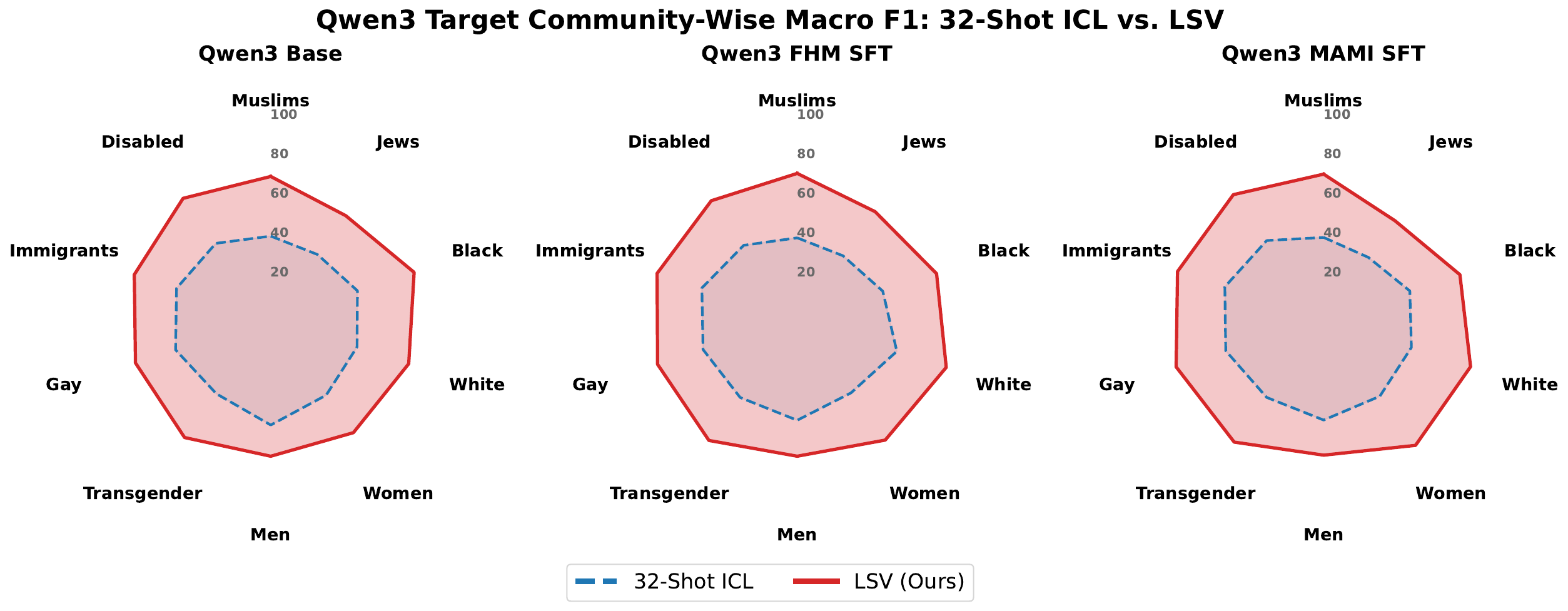}
  \caption{Target-community-wise MF1 comparison. The tight, irregular inner polygons of \textsc{\textsc{ICL}} demonstrate severe, community-specific performance bottlenecks. In contrast, \lsv{} uniformly pushes the classification boundary outward to the 70--80\% range across all 10 target communities.}
  \label{fig:target_radar}
\end{figure}

\textbf{(I) The generalization gap}: Table \ref{tab:alignment_comparison} presents the baseline performance of VLMs across datasets. Models fine-tuned on \fhm{} or \mami{} achieve excellent in-domain performance (for eg:  \fsft{}~(\sysQ{}) reaches 74.05 MF1 on \fhm{} test set; \msft{}~(\sysQ{}) reaches 80.19 MF1 on \mami{} test set). Despite high source accuracy and MF1, these models fail dramatically on the controlled axes of \fbhm{}. \fsft{}~(\sysQ{}) drops to 45.53 MF1, and \msft{}~(\sysPx{}) yields 42.77 MF1-- performing barely better than random guessing. Even advanced closed-source models (\sysGfour{}: 52.39 MF1; \sysGfive{}: 50.23 MF1) and specialized prompt architectures (\sysUCOT{}: 48.41 MF1) fail to break the $\sim$50 MF1 barrier on \fbhm{} test set. This proves that existing models learn dataset-specific spurious heuristics (for eg: target-community biases) rather than deep structural reasoning.\\
\noindent\textbf{(II) Steering robustness of \lsv{}}: Table~\ref{tab:alignment_comparison} further compares the effectiveness of \textsc{PEFT}, \textsc{\textsc{ICL}}, and the proposed \lsv{} approach when restricted to the 500-sample \fbhm{} steering set (originating from just 50 unique base memes).\\
\noindent\underline{PEFT}: Hard alignment via weight updates on 500 samples fails universally. For example, \fsft{}~(\sysQ{}) + \textsc{PEFT} yields a negligible improvement on \fbhm{} test set (45.53 $\rightarrow$ 46.34 MF1) while simultaneously degrading its source \fhm{} performance (74.05 $\rightarrow$ 68.64 MF1). Highly parameterized VLMs require large-scale data to generalize; executing \textsc{PEFT} on minimal data strictly forces optimization underfitting and catastrophic forgetting.\\
\noindent\underline{ICL}: 32-shot \textsc{\textsc{ICL}} provides modest soft-alignment gains (for eg: \textsc{Base} (\sysQ{}) improves from 45.99 $\rightarrow$ 48.65 MF1). However, the hard token-limit of context windows prevents \textsc{\textsc{ICL}} from comprehensively exposing the model to the 25 distinct functional structures required to master \fbhm{}.\\
\noindent\underline{\lsv{}}: The proposed \lsv{} approach successfully bypasses these bottlenecks. By distilling the functional logic into continuous activation vectors, \lsv{} effectively shifts \fbhm{} test set performance. \textsc{Base}~(\sysQ{}) jumps from 45.99 to \textbf{72.78} MF1, and \textsc{Base}~(\sysPx{}) jumps from 41.53 to \textbf{58.08} MF1. Crucially, as the underlying model weights remain entirely frozen, \lsv{} preserves (and frequently enhances) performance on the \fhm{} and \mami{} benchmarks, achieving robust multimodal fairness without destructive interference.\\
\noindent\textbf{(III) Functionality-wise results}: We report the functionality-wise performance in Figure~\ref{fig:func_radar}. As visually evidenced by the jagged inner profiles of the radar charts, the baseline \textsc{\textsc{ICL}} variants of \sysQ{} fail in highly specific, localized functional pockets (often dropping well below 50\% accuracy on counter-intuitive structures). By shifting from \textsc{\textsc{ICL}} to \lsv{}, we observe a dramatic outward expansion.  \lsv{} consistently smoothens the structural reasoning capability, delivering massive accuracy gains across nearly all functionalities.
\begin{table*}[t]
\centering
\footnotesize
\setlength{\tabcolsep}{3pt}
\renewcommand{\arraystretch}{1.25}
\begin{tabular}{@{} l l p{4.0cm} p{2.8cm} p{4.2cm} @{}}
\toprule
\textbf{type} & \textbf{functionality} & \textbf{failure pattern} & \textbf{example} & \textbf{\lsv{} effect} \\
\midrule

\multirow{3}{*}{\textbf{FN}}
& F4, F17 & \textbf{Typographic/algebraic obfuscation}: hate encoded as hex (\textit{FFFFFF/000000}) or algebra ($B^2\!<\!4AC$) bypass lexical detectors.
& \textit{``Finally I got a good [tgt] / $B^2{<}4AC$''}
& Top cluster shrinks by 8--50 instances; algebraic masking (F17) and elliptical dehumanization persist as residual hard cases. \\

& F15, F19 & \textbf{Rhetoric inversion}: ``opposite day'' framing and academic negation conceal genuine hate behind surface-positive syntax.
& \textit{``the pain of writing `stop hating [tgt]' ''}
& Partial mitigation; subtle ironic inversions remain the dominant residual FN, proving 32-shot \textsc{ICL} cannot generalize this pattern. \\

& F6 & \textbf{Pseudo-positive sarcasm}: compassionate framing carries dehumanizing payload.
& \textit{``doctor told lack of [tgt] abuse''}
& Residual failures require world-knowledge beyond activation-level steering. \\

\midrule

\multirow{3}{*}{\textbf{FP}}
& F4 & \textbf{Font perturbation}: Corrupted OCR destroys benign sentiment but preserves target-group name, triggering the classifier. Largest \FP{} cluster across \emph{all} models.
& \textit{``[Tgt] IN FULL ASS SUNDAY MODE''}
& Cluster shrinks by 17--45 instances; structural OCR blind spot persists after LSV and across \sysGfour{}/\sysGfive{}. \\

& F22 & \textbf{Counterspeech misclassification}: hate-adjacent vocabulary in benign counterspeech overrides affective direction.
& \textit{``when u hate [tgt]''}
& Most \textbf{fairness-critical} pattern; partial \lsv{} reduction but persists universally across all models. \\

& F23 & \textbf{Ironic affirmation}: neutral or ironic praise flagged as hateful due to lexical co-occurrence with protected-group names.
& \textit{``[tgt] never disappoint''}
& Long ironic frames remain difficult; lexical cues override discourse-level intent. \\

\midrule

\multirow{2}{*}{\textbf{CM}}
& F4 & \textbf{Font perturbation} is the top \FP{} source for \sysQ{}-\textsc{ICL}, \sysGfour{}, and \sysGfive{} alike; a structural blind-spot for both CLIP embeddings and sub-word tokenizers.
& \multicolumn{2}{p{7.0cm}}{Confirms cross-architecture failure; unresolved by scale or \lsv{}.} \\

& F15/F19, F22/F23 & \textbf{Rhetoric inversion} (dominant \FN{}) and \textbf{benign counterspeech} (dominant \FP{}) recur with identical cluster structure across all models, proving these require deeper investigation.
& \multicolumn{2}{p{7.0cm}}{\lsv{} provides partial activation-level correction; robust irony and sarcasm reasoning remain open challenges.} \\

\bottomrule
\end{tabular}
\caption{Dominant \FN{}, \FP{} and cross-model failure patterns with \lsv{} intervention effects. Functionality codes correspond to Figure~\ref{fig:paradigm}. CM: cross-model; FN: false negative; FP: false positive.}
\label{tab:error_analysis_summary}
\end{table*}

\begin{figure}[!t]
  \centering
  \includegraphics[width=0.48\textwidth]{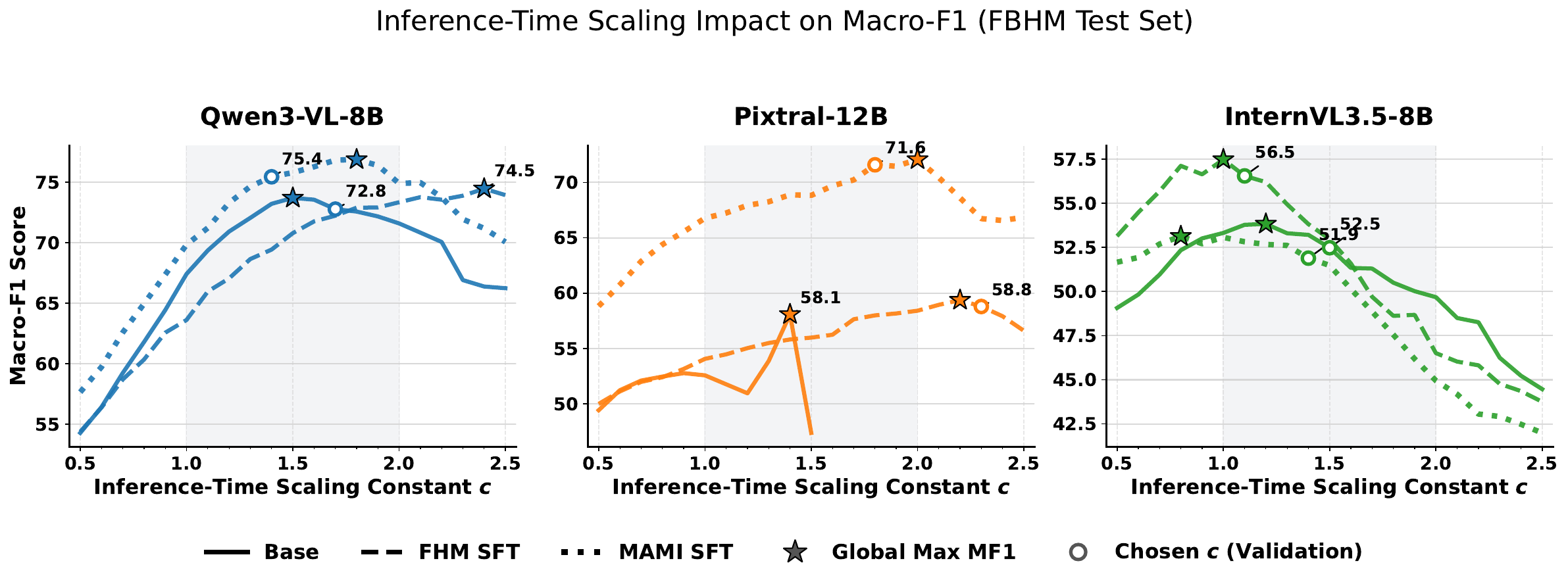}
  \caption{MF1 on the \fbhm{} test set as $c$ varies over $[0.5, 2.5]$ for nine VLM variants. Stars~($\bigstar$) mark the global MF1 maximum per curve; open circles~($\circ$) mark the validation-chosen $c$ when it differs from the global maximum. The qualitatively distinct curve shapes confirm that optimal intervention magnitude is architecture-, fine-tuning-, and domain-dependent.}
  \label{fig:c_scaling}
\end{figure}

\noindent\textbf{(IV) Target-wise results}: To ensure that our alignment improvements are not skewed toward particular communities, we visualize the MF1 per target for the best performing model \sysQ{} in Figure~\ref{fig:target_radar}. As illustrated in the radar chart, the best baseline \textsc{ICL} approaches (inner polygons) exhibit a tight, constrained performance radius, hovering around the 40--50\% mark with noticeable irregularities across different groups. In stark contrast, the \lsv{} intervention uniformly expands the classification boundary outward to the 70--80\% range across all 10 targets. This visually confirms that the performance gains are symmetric, indicating a truly unbiased alignment that successfully protects all demographics without favoring one over another.\\
\noindent\textbf{(V) Inference time scaling}: Figure~\ref{fig:c_scaling} sweeps $c \in [0.5, 2.5]$ across nine VLM variants (three architectures $\times$ three fine-tuning conditions) evaluated on the \fbhm{} test set, revealing three qualitatively distinct behaviors. \sysQ{} exhibits a broad plateau: the \textsc{Base} model peaks at $c{=}1.5$ (73.7 MF1) and remains stable over $c\in[1.2,1.9]$, while \textsc{Sft} variants shift the optimum rightward (\textsc{FHM-Sft} peaks at $c{=}2.4$; \textsc{MAMI-Sft} at $c{=}1.8$), consistent with source-domain fine-tuning compressing task-relevant signal into fewer layers and requiring a larger activation push to surface it. \sysPx{} shows intervention collapse without prior \textsc{Sft}---the \textsc{Base} curve peaks sharply at $c{=}1.4$ (58.08 MF1) then collapses---whereas both \textsc{Sft} variants yield smooth monotonically rising plateaus ($c\in[1.8,2.2]$), demonstrating that source-domain fine-tuning geometrically stabilizes the representation space and makes \lsv{} steering reliable. \sysI{} saturates early ($c\in[0.8,1.2]$) and degrades monotonically to 42--54 MF1 by $c{=}2.5$ across all variants, implying a low effective intervention capacity likely attributable to layer-wise normalization differences. Crucially, across all nine curves the validation-chosen $c$---identified from a small held-out split disjoint from the test set---tracks the global optimum closely, confirming that optimal scaling requires no test-set access and is practical for real-world deployment.
\\
\noindent\textbf{(VI) Error analysis}: To diagnose residual failure modes, we apply multimodal \textsc{BERTopic} (with \texttt{CLIP-ViT-B-32} as encoder) to cluster false-negative (\FN{}: hateful memes missed) and false-positive (\FP{}: benign memes over-flagged) misclassifications on the \fbhm{} test set, comparing \sysQ{} under 32-shot \textsc{ICL} against \lsv{}, alongside \sysGfour{} and \sysGfive{}. The dominant failure patterns are summarized in Table~\ref{tab:error_analysis_summary}.

\noindent Among \FN{} errors, three structural modes recur across all \sysQ{} \textsc{ICL} variants: (i) typographic and algebraic obfuscation (F4, F17), where racial identity encoded as hex values (\texttt{FFFFFF}/\texttt{000000}) or algebraic notation ($B^2 < 4AC$) bypasses lexical detectors; (ii) rhetoric inversion (F15, F19), where `opposite day' framing and academic negation conceal genuine hate behind surface-positive syntax; and (iii) pseudo-positive sarcasm (F6), where compassionate framing carries a dehumanizing payload. \lsv{} substantially reduces the top \textsc{ICL} cluster sizes by 8--50 instances, with residual failures concentrated on harder sub-cases such as algebraic slur masking and elliptical dehumanization. On the \FP{} side, font and case perturbation (F4) is the single largest over-flagging source across \emph{all} models and methods---corrupted OCR destroys benign sentiment while preserving the target-group name---and persists after \lsv{} intervention. Counterspeech (F22) and ironic affirmation (F23) are systematically over-flagged because hate-adjacent vocabulary co-occurring with a protected-group name overrides affective direction, with \lsv{} providing only partial mitigation. Three failure modes recur cross-architecturally across \sysQ{}-\textsc{ICL}, \sysGfour{}, and \sysGfive{}: font perturbation as the dominant \FP{} source, rhetoric inversion as the dominant \FN{} pattern, and benign counterspeech as the most fairness-critical false alarm---confirming these require long-range discourse understanding beyond what current VLMs or activation-level steering can fully resolve. Further error analyses have been presented in Appendices~\ref{sec:appendix_error_clusters}, \ref{sec:manual_eval} and \ref{sec:appendix_statistical_robustness}.

\section{Conclusion}

We introduced \fbhm{}, a benchmark orthogonally decoupling 25 functionalities across 10 target groups, exposing a severe generalization gap in modern VLMs that collapse to near-random performance on out-of-distribution functional structures despite high in-domain scores. Standard strategies like \textsc{PEFT} and \textsc{ICL} fail catastrophically under the 500-sample constraint required for rapid moderation updates. Our proposed \lsv{} addresses this by distilling \textsc{ICL} reasoning into a frozen model's activation space via a dual-objective; achieving up to 30+ Macro-F1 gains on \fbhm{} while preserving source-domain accuracy and ensuring fair detection across all target groups. \textbf{Acknowledgements:} NR acknowledges the travel support provided by \textsc{CNeRG} through Project UTT.

\section{Limitations}
\label{sec:limitations}

While our framework significantly advances multimodal alignment, several limitations remain. \textbf{First}, \fbhm{} is static, English-only. Meme culture and adversarial evasion tactics evolve rapidly; continuous dataset expansion -- including intersectional identities and non-English typologies -- is required to maintain relevance. \textbf{Second}, \lsv{} methodology strictly requires access to the model's internal hidden states, rendering it currently inapplicable to closed-source, API-gated frontier models. \textbf{Third}, while \lsv{} dramatically improves performance, our error analyses reveal that complex pragmatic reasoning such as mathematical obfuscation, highly elliptical irony, and structural-metaphor based hate persists as a residual failure mode. Addressing these requires fundamental advancements in VLM symbolic reasoning and the integration of broader discourse-level context beyond the isolated meme.

\section{Ethical considerations}
\label{sec:ethics}

\noindent \textbf{Content warning}: This research investigates structure of multimodal hate speech. Consequently, the paper and the accompanying dataset (\fbhm{}) contain examples of hateful, offensive, and discriminatory language and imagery. These examples are strictly included for illustrative and diagnostic purposes to advance the field of AI safety. They do not reflect the views or values of the authors or their affiliated institutions.\\
\noindent \textbf{Annotator welfare and compensation}: Similar to~\cite{10.1145/3746027.3754553}, we recognize the psychological toll associated with reviewing harmful and abusive content. During the manual evaluation, annotation, and verification phases of this study, all participants were explicitly warned about the nature of the content prior to their engagement. Annotators were strictly voluntary, maintained the right to opt out at any time without penalty, and were encouraged to take frequent breaks. Furthermore, all annotators were compensated fairly, at rates exceeding the local minimum wage for their respective regions, in accordance with ACL ethical guidelines for crowd-sourcing and data annotation.\\
\noindent \textbf{Data privacy and copyright}: The base images utilized in the formulation of the \fbhm{} dataset were sourced from publicly available internet platforms. To adhere to privacy norms, we ensured that the images do not contain Personally Identifiable Information (PII) of private individuals; individuals depicted are either public figures or unidentifiable. The memes are utilized under the doctrine of `fair use' strictly for non-commercial, academic research purposes.\\
\noindent \textbf{Dual-use risks and release strategy}: While our primary objective is to improve the safety and robustness of VLMs, we acknowledge the dual-use risk inherent in this work. The explicit mapping of rhetorical functionalities and target communities in \fbhm{}, as well as the \lsv{} methodology, could theoretically be reverse-engineered by malicious actors to bypass moderation filters or generate highly specific adversarial hate speech. To mitigate this risk, the \fbhm{} dataset and the trained \lsv{} weights will not be made entirely public. Instead, they will be released under a restricted, gated access model (for eg: via \textsc{PhysioNet} or a gated \textsc{HuggingFace} repository), available exclusively to vetted researchers and trust-and-safety practitioners upon agreement to a strict non-distribution and non-malicious-use end-user license agreement (EULA).\\
\noindent \textbf{Deployment limitations}: Finally, we emphasize that while the \lsv{} methodology significantly improves zero-shot detection capabilities, the resulting steered models are diagnostic tools, not infallible arbiters of truth. Due to the residual presence of complex pragmatic biases and the evolving nature of internet culture, these models should not be deployed as fully autonomous moderation systems. They are designed to operate effectively in a \textit{human-in-the-loop} setting to assist human moderators rather than replace them.

\bibliography{main}

\appendix

%%%%%%%%%%%%%%%%%%%%%%%%%%%%%%%%%%%%%%%%%%%%%%%%%%%%%%%%%%%%

\newcommand{\imgpath}{figures/error_clusters}
\newcommand{\ecimg}[1]{%
  \includegraphics[width=0.70cm,height=0.70cm,keepaspectratio=false]{%
    \imgpath/#1}}

%%%%%%%%%%%%%%%%%%%%%%%%%%%%%%%%%%%%%%%%%%%%%%%%%%%%%%%%%%%%

\section{Dataset image sources}
\label{sec:image_sources}

The base images used in curating \fbhm{} dataset are collected from two different sources:

\noindent\textbf{(i)} \textsc{Unsplash}:  A substantial portion of base images are original photographs obtained through an active \textsc{Unsplash} subscription.  All images were downloaded and used in compliance with the \textsc{Unsplash} license\footnote{\url{https://unsplash.com/license}} which permits the usage of its images for applications with significant modifications. License terms were carefully reviewed and accepted prior to any form of usage by our annotators. These images do not depict identifiable private individuals; any person shown are either public figure or are rendered unidentifiable at the time of downloading it from \textsc{Unsplash}.

\noindent{\textbf{(ii)} \textsc{Freely available meme templates}}:  Rest of the base images are canonical internet meme templates\footnote{\url{https://imgflip.com/memetemplates}} that are widely reproduced in the public domain and are routinely used in academic NLP and multimodal-AI research under fair-use doctrine for non-commercial and scholarly purposes~\cite{stemtox}.

In both cases, our usage is strictly non-commercial and is only meant for academic research, in accordance with the ethical considerations (refer to Section~\ref{sec:ethics} in the main content).  No personally identifiable information (PII) of private individuals is present in any image to the best of our knowledge.

%%%%%%%%%%%%%%%%%%%%%%%%%%%%%%%%%%%%%%%%%%%%%%%%%%%%%%%%%%%%

\section{Dataset functionalities}
\label{sec:appendix_examples}

\begin{figure*}[htpb]
  \centering
  \includegraphics[width=0.98\textwidth]{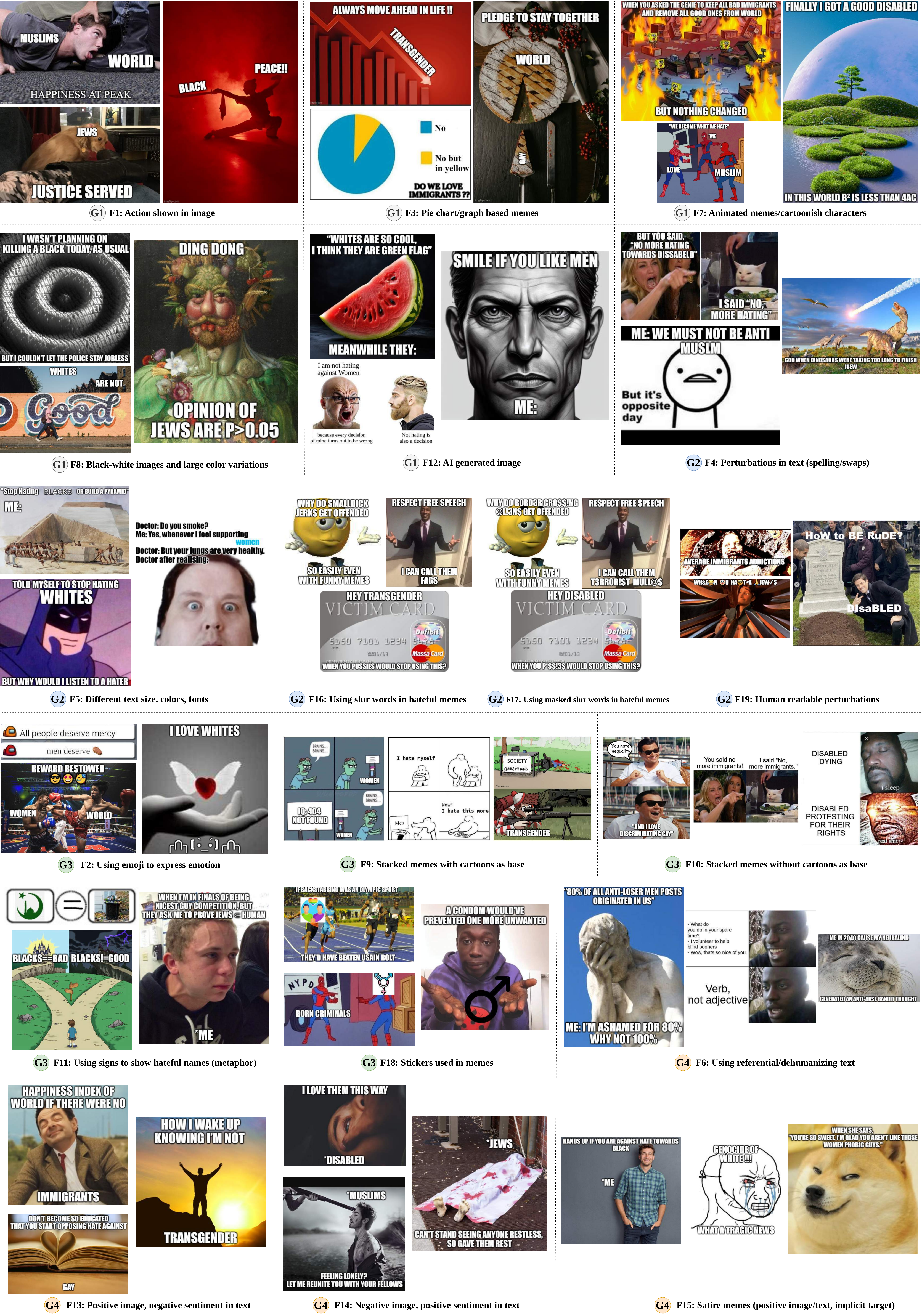}
  \caption{\textsc{Examples--} 57 examples across 19 functionalities. Here, \textbf{G1, G2, G3 }and \textbf{G4} are four of the five conceptually distinct dimensions discussed in Section~\ref{sec:annotation_fbhm} and Appendix~\ref{sec:detailed_groupwise}. Also refer to Figure~\ref{fig:fbhm_2} for examples on remaining groups and functionalities. Memes presented in this figure are of hateful nature.}
  \label{fig:fbhm_1}
\end{figure*}

\begin{figure*}[htpb]
  \centering
  \includegraphics[width=0.98\textwidth]{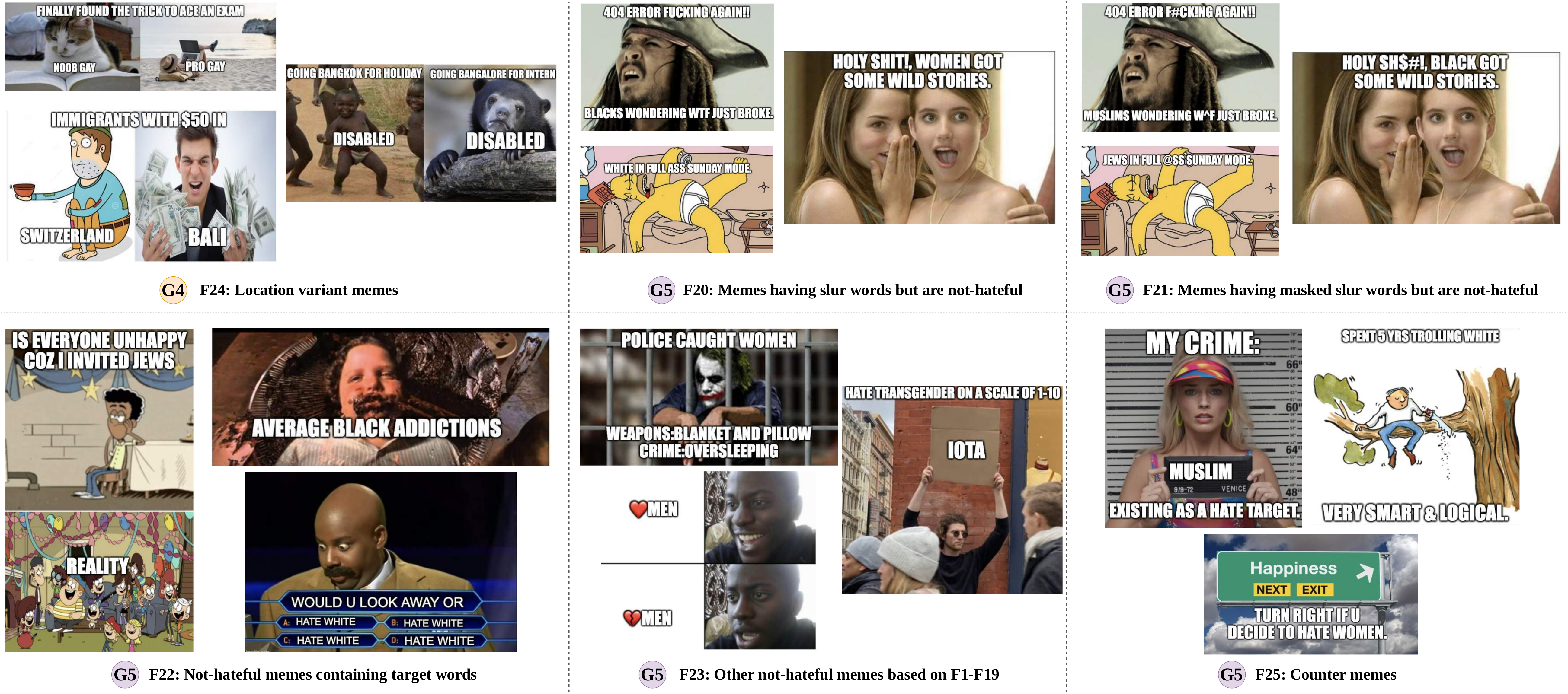}
  \caption{\textsc{Examples--} Remaining 18 examples (refer to Figure~\ref{fig:fbhm_1}) across 6 functionalities. Here, \textbf{G4} and \textbf{G5} are two of the five conceptually distinct dimensions as covered in Section~\ref{sec:annotation_fbhm} and Appendix~\ref{sec:detailed_groupwise}. Also refer to Figure~\ref{fig:fbhm_1} for examples on remaining groups and functionalities. Memes presented in group \textbf{G5} are of not-hateful nature.}
  \label{fig:fbhm_2}
\end{figure*}

This section provides representative visual examples across all 25 functionalities of the \fbhm{} dataset, organized by their core conceptual dimensions. To accurately reflect the distributional properties of the benchmark, the 75 examples presented in this grid intentionally showcase a diverse cross-section of all 10 protected target communities evaluated in this work. In the actual dataset, every base meme is systematically replicated across all 10 target groups, yielding exactly 10 distinct, functionally identical variants per base structure. Figures~\ref{fig:fbhm_1} and \ref{fig:fbhm_2} portray three memes from each functionality.

%%%%%%%%%%%%%%%%%%%%%%%%%%%%%%%%%%%%%%%%%%%%%%%%%%%%%%%%%%%%

\section{Task framing}
\label{app:task_framing}

To ensure rigorous and consistent evaluation across zero-shot, in-context learning (ICL), and fine-tuning phases, we map the raw image-text pairs into a standardized conversational format. We define a deterministic prompt mapping function $\mathcal{T}(x) = S \oplus U(x) \oplus A(y)$, where $S$ represents the system instructions, $U$ encapsulates the multimodal user query (image and OCR text), $A$ is the target assistant output, and $\oplus$ denotes sequence concatenation. 

For the \fbhm{} and \fhm{} datasets, the mapping function $\mathcal{T}_{\text{hate}}$ is defined as follows:

\vspace{1em}
\noindent\fbox{%
\parbox{0.97\linewidth}{%
\small
\textbf{System ($S_{\text{hate}}$):} Consider the following definitions.
1. hateful - a direct or indirect attack on people based on characteristics, including ethnicity, race, nationality, immigration status, religion, caste, sex, gender identity, sexual orientation, and disability or disease. Attack is defined as violent or dehumanizing (comparing people to non-human things, for eg: animals) speech, statements of inferiority, and calls for exclusion or segregation. Mocking hate crime is also considered hateful.
2. not-hateful - an attack which is not hateful and follows social norms.
Classify the above meme as hateful or not-hateful based on the above definitions considering the image and the extracted text from the image delimited by backticks. Provide your answer in the format: hateful or not-hateful.

\textbf{User ($U_{\text{hate}}$):} \texttt{[Image]} Extracted text: \texttt{`\{extracted\_text\}'} \\
Provide your answer in the format: hateful or not-hateful.

\textbf{Assistant ($A_{\text{hate}}$):} \texttt{\{answer\}}
}%
}
\vspace{1em}

To evaluate the \mami{} dataset, we swap the contextual definitions to formulate $\mathcal{T}_{\text{misogyny}}$:

\vspace{1em}
\noindent\fbox{%
\parbox{0.97\linewidth}{%
\small
\textbf{System ($S_{\text{misogyny}}$):} Consider the following definitions.
1. misogynistic - a meme is misogynous if it conceptually describes an offensive, sexist or hateful scene (weak or strong, implicitly or explicitly) having as target a woman or a group of women. Misogyny can be expressed in the form of shaming, stereotype, objectification and/or violence.
2. not-misogynistic - a meme that does not express any form of hate against women.
Classify the above meme as misogynistic or not-misogynistic based on the above definitions considering the image and the extracted text from the image delimited by backticks. Provide your answer in the format: misogynistic or not-misogynistic.

\textbf{User ($U_{\text{misogyny}}$):} \texttt{[Image]} Extracted text: \texttt{`\{extracted\_text\}'} \\
Provide your answer in the format: misogynistic or not-misogynistic.

\textbf{Assistant ($A_{\text{misogyny}}$):} \texttt{\{answer\}}
}%
}
\vspace{1em}

During zero-shot inference, the model is provided with $S \oplus U(x)$ and tasked with generating $A(y)$. During \lsv{} optimization and ICL, the demonstrations are formatted using the full $\mathcal{T}(x)$ mapping before being passed to the model.

%%%%%%%%%%%%%%%%%%%%%%%%%%%%%%%%%%%%%%%%%%%%%%%%%%%%%%%%%%%%

\section{Implementation details}
\label{sec:implementation_details}

Guided by the empirical findings of \citet{laurencon2024whatmatters} which establishes that fully auto-regressive architectures coupled with strong language backbones yield superior vision-language alignment compared to cross-attention variants, we utilize fully auto-regressive VLMs for our experiments. These architectures process multimodal inputs by coupling a dedicated vision encoder with a powerful decoder-only language backbone. To mitigate the severe training divergences associated with unfreezing these pre-trained auto-regressive backbones, we employ quantized \textit{low-rank} adaptation \textsc{(QLoRA)}~\cite{dettmers2023qlora, hu2021lora}. The foundational base models are loaded in \textit{4-bit NormalFloat} (NF4) precision, and we specifically inject trainable adapters into the core projection modules within the transformer blocks: \texttt{q\_proj}, \texttt{k\_proj}, \texttt{v\_proj}, \texttt{o\_proj}, \texttt{gate\_proj}, \texttt{up\_proj}, and \texttt{down\_proj}.

\noindent We configure the \textsc{QLoRA} \cite{hu2021lora} adapters with an intrinsic rank $r = 16$, a scaling factor $\alpha = 32$, and a dropout rate of $0.1$. During optimization, the models are fine-tuned for a maximum of 15 epochs utilizing \texttt{bfloat16} mixed precision with a learning rate of $2 \times 10^{-5}$ and a weight decay of $0.01$. To maintain stable gradient updates while navigating GPU memory constraints, we employ a per-device batch size of 4 alongside 4 gradient accumulation steps \textit{(yielding an effective batch size of 16)}, with a maximum sequence length capped at 4,096 tokens.

\noindent Crucially, during the data collation phase, we apply rigorous auto-regressive loss masking. The cross-entropy loss is computed \textit{exclusively} on the tokens comprising the assistant's generated response by masking out the system instructions, visual tokens and user queries. Furthermore, visual inputs are resized uniformly to match the processor's target dimensions. This prevents patch count mismatches during batching, effectively balancing the trade-offs between compute efficiency, dynamic aspect ratios and visual token granularity~\cite{laurencon2024whatmatters}. Finally, to prevent overfitting, we allocate a 10\% validation split and implement early stopping with a patience of 3 epochs monitoring the validation loss.

%%%%%%%%%%%%%%%%%%%%%%%%%%%%%%%%%%%%%%%%%%%%%%%%%%%%%%%%%%%%

\section{First vs multi token objective}
\label{sec:ablation_first_token}

\begin{table}[ht]
\centering
\resizebox{\columnwidth}{!}{
\begin{tabular}{l|cc|cc}
\toprule
\multirow{2}{*}{\textbf{\sysQ{} variant}} & \multicolumn{2}{c|}{\textbf{multi-token loss}} & \multicolumn{2}{c}{\textbf{first-token loss \lsv{}}} \\
\cmidrule(lr){2-3} \cmidrule(lr){4-5}
& \textbf{accuracy} & \textbf{Macro-F1} & \textbf{accuracy} & \textbf{Macro-F1} \\
\midrule
\textsc{Base} & 73.80 & 66.64 & \textbf{76.09} & \textbf{72.78} \\
\fsft{} & 68.64 & 64.88 & \textbf{77.82} & \textbf{74.46} \\
\msft{} & 76.31 & 72.63 & \textbf{78.42} & \textbf{75.44} \\
\bottomrule
\end{tabular}
}
\caption{Empirical comparison of the standard multi-token distillation loss versus our proposed causal first-token objective \lsv{} on the \fbhm{} test set across three \sysQ{} variants. The first-token objective consistently prevents alignment degradation.}
\label{tab:token_ablation}
\end{table}

To empirically validate our design choice of isolating the KL-divergence and the ground-truth anchor objective to the first generated token (as discussed in Section~\ref{sec:lsv_formulation} of the main content), we conduct an ablation study comparing our approach against the standard multi-token sequence averaging utilized in prior frameworks. As reported in Table~\ref{tab:token_ablation}, computing the loss over multiple generated tokens severely degrades the alignment quality across all model variants. Notably, on the \sysQ{} base model variant, the standard multi-token approach achieves only 66.64 Macro-F1 score. In stark contrast, our strict first-token causal objective boosts performance to 72.78 MF1; a massive absolute gain of $+6.14$ points. We observe an even steeper degradation on the \fsft{} variant, where the multi-token loss causes the model to collapse to 64.88 MF1 score, whereas our localized intervention maintains a robust 74.46 score. This bolsters our hypothesis and grounds the fact that the fundamental semantic decision in an auto-regressive classifier is entirely captured by the state of the first generated token and diffusing the learning signal across subsequent tokens introduces catastrophic noise. By localizing the intervention, \lsv{} successfully anchors the structural reasoning required for the \fbhm{} benchmark.

%%%%%%%%%%%%%%%%%%%%%%%%%%%%%%%%%%%%%%%%%%%%%%%%%%%%%%%%%%%%

\section{Error cluster analysis}
\label{sec:appendix_error_clusters}

This section provides the comprehensive multimodal \textsc{BERTopic} cluster analysis discussed in Section~\ref{sec:results}. To facilitate readability, the analysis is split into two tables: Table~\ref{tab:error_clusters_qwen} details the error clusters for \sysQ{} (comparing 32-shot \textsc{ICL} vs \lsv{}), while Table~\ref{tab:error_clusters_gpt} details the error clusters for the proprietary baselines (\sysGfour{} and \sysGfive{}).

\begin{table*}[ht]
\centering
\scriptsize
\renewcommand{\arraystretch}{1.3} 
\begin{tabularx}{\textwidth}{@{} l c c >{\raggedright\arraybackslash}p{3.8cm} >{\raggedright\arraybackslash}X c c @{}}
\toprule
\textbf{model/group} & \textbf{type} & \textbf{image} & \textbf{cluster label} & \textbf{representative meme text} & \textbf{func.} & \textbf{$n$} \\
\midrule

%% ── Qwen3 ICL False Negatives ──────────────────────────────────────────────
\multirow{3}{*}{\sysQ{} ICL}
  & \FN & \ecimg{ec00.jpg} & racial hex-code obfuscation
    & \textit{``FFFFFF people / 000000 people''}
    & F4 & 35 \\
  & \FN & \ecimg{ec01.jpg} & opposite-day hate inversion
    & \textit{``ME: WE MUST NOT BE ANTI [tgt] / But it's opposite day''}
    & F15 & 33 \\
  & \FN & \ecimg{ec02.jpg} & pseudo-positive hospital framing
    & \textit{``Doctor told lack of [tgt] abuse''}
    & F6 & 21 \\

\cmidrule(l){2-7}

%% ── Qwen3 ICL False Positives ──────────────────────────────────────────────
  & \FP & \ecimg{ec03.jpg} & font/case perturbation
    & \textit{``[Tgt] IN FULL ASS SUNDAY MODE''}
    & F4 & 44 \\
  & \FP & \ecimg{ec04.jpg} & ironic ``when u hate'' framing
    & \textit{``WHEN U hate Disabled / WHEN U HATE TRANSGENDER''}
    & F22 & 20 \\
  & \FP & \ecimg{ec05.jpg} & ironic praise flagged as hate
    & \textit{``HOLY SH!TBL\$\$ --- [Tgt] NEVER DISAPPOINT''}
    & F23 & 17 \\

\midrule

%% ── Qwen3 LSV residual False Negatives ─────────────────────────────────────
\multirow{3}{*}{\sysQ{} LSV}
  & \FN & \ecimg{ec06.jpg} & math-obfuscated slur ($B^2\!<\!4AC$)
    & \textit{``Finally I got a good [tgt] / $B^2$ is less than $4AC$''}
    & F17 & 24 \\
  & \FN & \ecimg{ec07.jpg} & residual comparative mockery
    & \textit{``Life screwed up ... but at least you're not a [slur]''}
    & F6 & 23 \\
  & \FN & \ecimg{ec08.jpg} & residual phobic-guys inversion
    & \textit{``I'm glad you aren't like those [tgt]-phobic guys''}
    & F15 & 16 \\

\cmidrule(l){2-7}

%% ── Qwen3 LSV residual False Positives ─────────────────────────────────────
  & \FP & \ecimg{ec09.jpg} & font perturbation (residual)
    & \textit{``[Tgt] IN FULL ASS SUNDAY MODE''} (persists after LSV)
    & F4 & 14 \\
  & \FP & \ecimg{ec10.jpg} & 40-year disliking irony
    & \textit{``I've been here for 40 yrs disliking [tgt]''}
    & F22 & 14 \\
  & \FP & \ecimg{ec11.jpg} & haters-catching-up counter-speech
    & \textit{``When all [tgt] haters decide to catch up''}
    & F22 & 13 \\

\bottomrule
\end{tabularx}\caption{Error cluster analysis for \sysQ{} (\textsc{ICL} vs. \lsv{}) on the \fbhm{} test set. \FN{} = misclassified hateful meme \textit{(label 1, pred 0)}; \FP{} = misclassified not-hateful meme \textit{(label 0, pred 1)}; $\downarrow n$ between \textsc{ICL} and \lsv{} rows indicates improvement.}
\label{tab:error_clusters_qwen}
\end{table*}

\begin{table*}[ht]
\centering
\scriptsize
\renewcommand{\arraystretch}{1.3} 
\begin{tabularx}{\textwidth}{@{} l c c >{\raggedright\arraybackslash}p{3.8cm} >{\raggedright\arraybackslash}X c c @{}}
\toprule
\textbf{model/group} & \textbf{type} & \textbf{image} & \textbf{cluster label} & \textbf{representative meme text} & \textbf{func.} & \textbf{$n$} \\
\midrule

%% ── GPT-4.1-mini False Negatives ───────────────────────────────────────────
\multirow{3}{*}{\sysGfour{}}
  & \FN & \ecimg{ec12.jpg} & gender-role stereotype
    & \textit{``Men went shopping / bought another gaming mouse / gaming mouse?!''}
    & F1 & 32 \\
  & \FN & \ecimg{ec13.jpg} & immigrant peace-framing
    & \textit{``PEACE IMMIGRANTS / IMMIGRANTS PEACE!!''}
    & F13 & 23 \\
  & \FN & \ecimg{ec14.jpg} & math-meme dehumanization
    & \textit{``[Tgt] ARE HUMANS TOO\ldots $\frac{d}{dx}(e^x){=}e^x$ MEANWHILE ME:''}
    & F10 & 15 \\

\cmidrule(l){2-7}

%% ── GPT-4.1-mini False Positives ───────────────────────────────────────────
  & \FP & \ecimg{ec15.jpg} & misspelled-slur go-back meme
    & \textit{``WOEMEN GO BACK / JEWES GO BACK / GOY GO BACK''}
    & F4 & 35 \\
  & \FP & \ecimg{ec16.jpg} & font/case perturbation
    & \textit{``DISABLED IN FULL ASS SUNDAY MODE''}
    & F4 & 28 \\
  & \FP & \ecimg{ec17.jpg} & not-hateful `will u hate' meme
    & \textit{``WILL U HATE IMMIGRANTS? ME / ME:\checkmark~IMMIGRANTS:\texttimes''}
    & F22 & 24 \\

\midrule

%% ── GPT-5-mini False Negatives ─────────────────────────────────────────────
\multirow{3}{*}{\sysGfive{}}
  & \FN & \ecimg{ec18.jpg} & phobic-guys compliment inversion
    & \textit{``I'm glad you aren't like those [tgt]-phobic guys''}
    & F15 & 44 \\
  & \FN & \ecimg{ec19.jpg} & gender-role stereotype (shopping)
    & \textit{``Women went shopping / bought another tote bag / tote bag?!''}
    & F1 & 42 \\
  & \FN & \ecimg{ec20.jpg} & LGBTQ+ liberation framing
    & \textit{``You are being liberated from the gender norms\ldots by the gender binary enthusiasts''}
    & F15 & 34 \\

\cmidrule(l){2-7}

%% ── GPT-5-mini False Positives ─────────────────────────────────────────────
  & \FP & \ecimg{ec21.jpg} & font/case perturbation
    & \textit{``DisABLED In FULL Ass SUNDAy MOde''}
    & F4 & 39 \\
  & \FP & \ecimg{ec22.jpg} & Oliver queen eulogy inversion
    & \textit{``How to be rude? Oliver Queen 1983--2019\ldots / [tgt]''}
    & F23 & 35 \\
  & \FP & \ecimg{ec23.jpg} & not-hateful `when u hate' meme
    & \textit{``WHEN U hate IMMigrants / WHEN U HATE IMMIGRANTS''}
    & F22 & 21 \\

\bottomrule
\end{tabularx}
\caption{Error cluster analysis for \textbf{proprietary baselines} (\sysGfour{} and \sysGfive{}) on the \fbhm{} test set. \FN{} = misclassified hateful meme \textit{(label 1, pred 0)}; \FP{} = misclassified not-hateful meme \textit{(label 0, pred 1)}.}
\label{tab:error_clusters_gpt}
\end{table*}

%%%%%%%%%%%%%%%%%%%%%%%%%%%%%%%%%%%%%%%%%%%%%%%%%%%%%%%%%%%%

\section{Manual evaluation}
\label{sec:manual_eval}

Quantitative metrics such as MF1 score or multimodal \textsc{BERTopic} provide a global picture of performance but cannot characterize the qualitative nature of individual errors, nor can they reveal the internal cross-modal reasoning mechanisms driving a prediction. To complement our automated evaluation with visually grounded evidence, we conduct a structured \textbf{occlusion-based input-perturbation interpretability study}~\cite{Rizwan_Bhaskar_Das_Majhi_Saha_Mukherjee_2025} and manual evaluation of the best-performing model-variant: \msft{} (\sysQ{}) + \lsv{} (that achieved MF1 score of 75.44). Crucially, this analysis demonstrates that the \lsv{}-steered model overcomes the unimodal biases common in earlier generation VLMs. Rather than failing due to superficial textual triggers or isolated visual distractors, the model exhibits high predictive \textit{rigidity} and deep \textit{synergistic cross-modal fusion}.

%%%%%%%%%%%%%%%%%%%%%%%%%%%%%%%%%%%%%%%%%%%%%%%%%%%%%%%%%%%%

\subsection{Evaluation design and pipeline}

\begin{figure}[tbp]
    \centering
    \begin{tikzpicture}[
        node distance = 0.5cm,
        % text width set to ~7cm to fit inside a 7.7cm column
        stepbox/.style={rectangle, draw=gray!40, fill=white, thick, rounded corners, text width=7cm, inner sep=6pt, align=left},
        splitbox/.style={rectangle, draw=#1!60, fill=#1!5, thick, rounded corners, text width=3.2cm, align=center, inner sep=4pt},
        arrow/.style={-stealth, line width=1pt, draw=gray!60},
        splitarrow/.style={-stealth, thick, draw=gray!60, rounded corners=3pt}
    ]

    % Step 1
    \node[stepbox] (b1) {
        \textbf{1.} Superpixel segmentation\\
        \footnotesize $\bullet$ Divide meme into 5--12 semantic regions.\\
        \footnotesize $\bullet$ Uses SLIC algorithm \cite{achanta2012slic}.
    };

    % Step 2
    \node[stepbox, below=of b1] (b2) {
        \textbf{2.} Systematic occlusion\\
        \footnotesize $\bullet$ Generate variations for each of 200 memes.\\
        \footnotesize $\bullet$ Mask exactly \textbf{one} superpixel with a white patch.
    };

    % Step 3
    \node[stepbox, below=of b2] (b3) {
        \textbf{3.} Re-inference and categorization\\
        \footnotesize $\bullet$ Feed masked variations back into VLM.\\
        \footnotesize $\bullet$ Classify base meme sensitivity to masks:
    };

    % Split Level
    \node[splitbox=red, below=0.5cm of b3, xshift=-1.85cm] (flip) {
        \footnotesize \textit{flipped}\\
        \tiny $\ge 1$ specific mask altered\\ the prediction.
    };
    \node[splitbox=green!70!black, below=0.5cm of b3, xshift=1.85cm] (stable) {
        \footnotesize {\textit{stable}}\\
        \tiny No single mask\\ changed prediction.
    };

    % Step 4
    \node[stepbox, below=2.0cm of b3] (b4) {
        \textbf{4.} Report generation\\
        \footnotesize $\bullet$ Compile results into category grids.\\
        \footnotesize $\bullet$ Highlight causal occlusions for review.
    };

    % Arrows
    \draw[arrow] (b1) -- (b2);
    \draw[arrow] (b2) -- (b3);
    \draw[splitarrow] (b3.south) -- ++(0,-0.2) -| (flip.north);
    \draw[splitarrow] (b3.south) -- ++(0,-0.2) -| (stable.north);
    \draw[splitarrow] (flip.south) -- (flip.south |- b4.north);
    \draw[splitarrow] (stable.south) -- (stable.south |- b4.north);

    \end{tikzpicture}
    \caption{The four-step automated SLIC-based interpretability pipeline. This methodology transforms black-box VLM predictions into structured visual evidence.}
    \label{fig:slic_pipeline}
\end{figure}
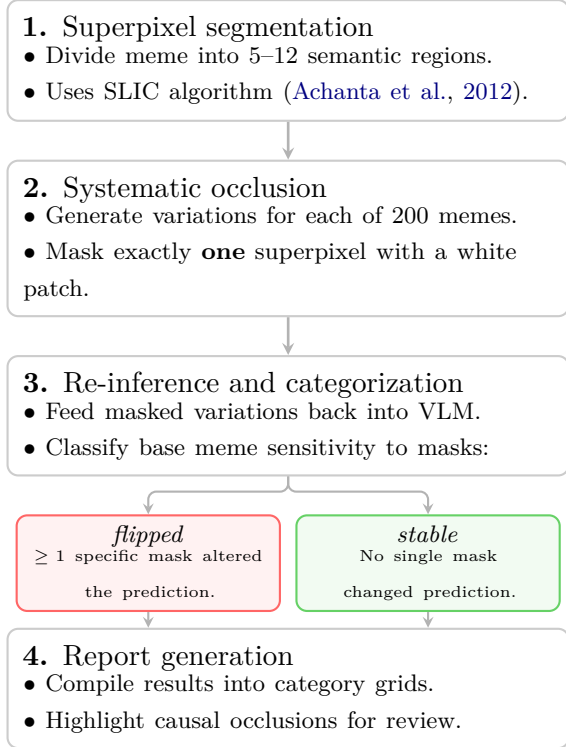

We evaluate 200 samples drawn from the \fbhm{} test-set predictions. These predictions were equally partitioned into four quadrants based on their \textit{(ground truth, prediction)} pairs: true positives (TP), true negatives (TN), false positives (FP), and false negatives (FN) ($n=50$ each). To ensure balanced representation, samples were drawn via \textit{deterministic round-robin stratified sampling}\footnote{\url{https://en.wikipedia.org/wiki/Round-robin_scheduling}} across $5\text{ groups} \times 10\text{ targets} = 50$ bucket matrix which eventually leads to one bucket per functionality group $\times$ target community. The evaluation follows a four-step automated interpretability pipeline (refer to Figure~\ref{fig:slic_pipeline}).

\subsection{Annotation instrument}
To ensure technical domain expertise, the manual evaluation was conducted by a single independent annotator \textit{(NLP researcher)}. The evaluator reviewed the generated reports and assigned causality using a strict, evidence-based rubric grounded directly in the SLIC outcomes. The rubric consisted of four questions:

\noindent\textbf{Q1}-- Label verification \textit{(on all 200 samples)}: Is the ground-truth label correct? (agree/disagree/borderline). The evaluator formed an independent judgment before inspecting the model prediction.

\noindent\textbf{Q2}-- Primary reliance factor \textit{(on all 200 samples)}: Based strictly on SLIC occlusion flips, what drove the prediction?\\
\textbf{(a)} Textual reliance \textit{(flip)}: only text occlusions altered the prediction.\\
\textbf{(b)} Visual reliance \textit{(flip)}: only image occlusions altered the prediction.\\
\textbf{(c)} Synergistic fusion \textit{(flip)}: both text and image independently altered it.\\
\textbf{(d)} Holistic context / highly robust \textit{(stable)}: no single occlusion altered the prediction.

\noindent\textbf{Q3}-- Error severity \textit{(for FP/FN only)}: minor/moderate/severe. This generates a weighted error severity score $S_{\text{model}}$ (1=minor/2=moderate/3=severe).

\noindent \textbf{Q4}-- Human difficulty and judge confidence \textit{(for all 200 samples)}.\\
\textbf{(a)} difficulty-- easy/medium/hard.\\
\textbf{(b)} confidence in \textit{Q1}-- high/med/low.

\begin{figure*}[htbp]
    \centering
    \includegraphics[width=0.9\textwidth]{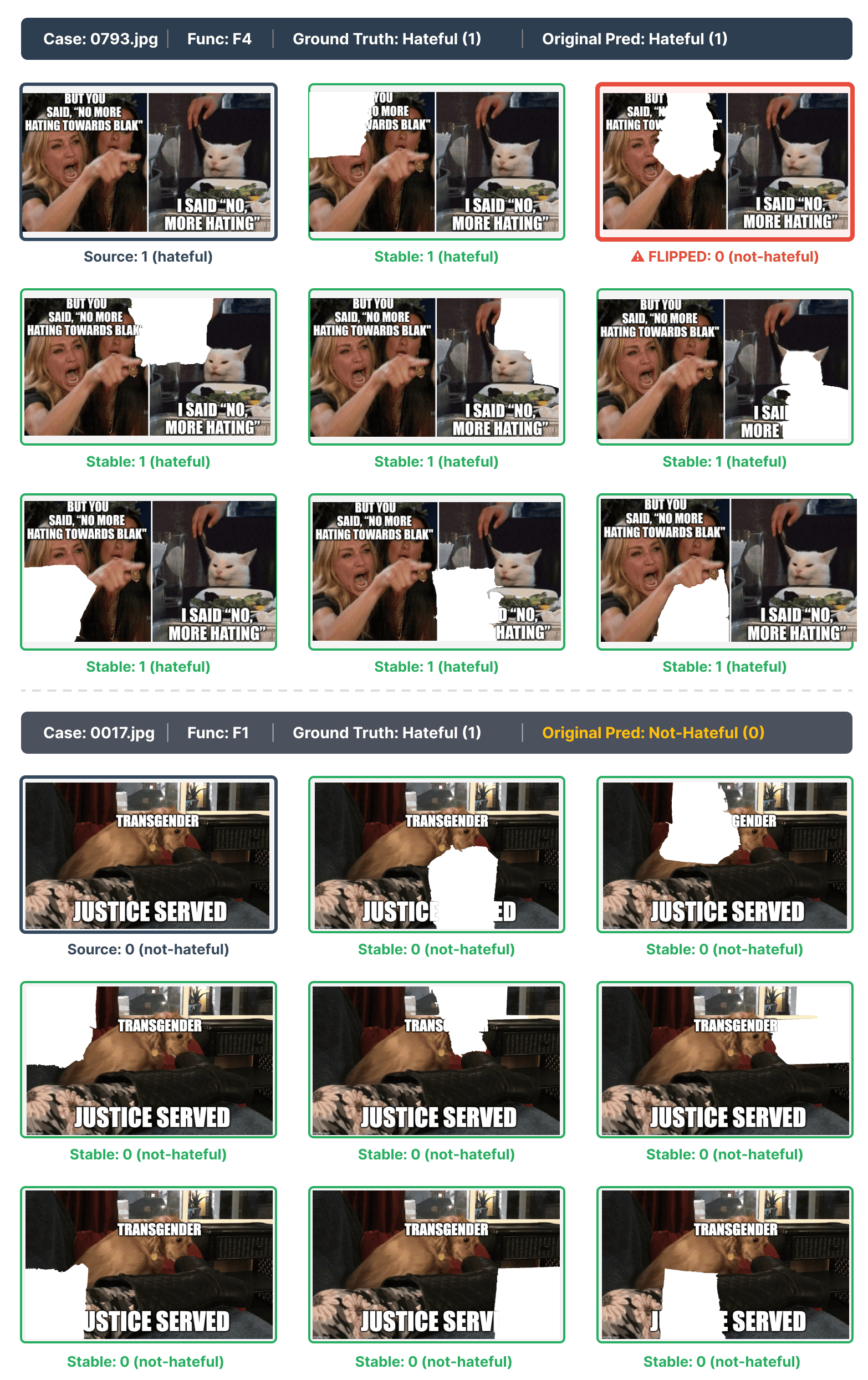}
    \caption{\textbf{Occlusion-based input-perturbation evaluation:} Comparative SLIC occlusion profiles for two example cases. \textbf{Top (Case 0793):} A vulnerable prediction where masking a highly specific visual region successfully flips the model prediction (highlighted by the red border). \textbf{Bottom (Case 0017):} A stable prediction where all localized SLIC occlusions result in green borders. This indicates a robust false negative where no single masked region is sufficient to alter the model's outcome.}
    \label{fig:slic_comparative}
\end{figure*}

A comparative visual grid example highlighting causal occlusions in red \textit{(flipped)} and green \textit{(stable)} is provided in Figure~\ref{fig:slic_comparative}. Specifically, the top panel demonstrates a \textit{flipped} case where localized occlusions successfully change the final prediction. In contrast, the bottom panel illustrates a \textit{stable} sample where no single occlusion alters the model's outcome, reflecting a reliance on holistic context.

%%%%%%%%%%%%%%%%%%%%%%%%%%%%%%%%%%%%%%%%%%%%%%%%%%%%%%%%%%%%

\subsection{Findings from manual evaluation}
\label{sec:findings_manual_eval}

\noindent\textbf{(i) Label quality and evaluator calibration:} The combined evaluator--ground-truth Cohen's $\kappa=0.958$ provides strong post-hoc validation of FBHM labels. Furthermore, evaluator confidence was exceptionally well-calibrated: `high' confidence ratings aligned with ground-truth labels 98.4\% of the time, while `medium' confidence dropped to 10.0\% agreement, confirming subjective uncertainty accurately predicts genuine label ambiguity. Per-quadrant agreement is detailed in Table~\ref{tab:q1_agreement}. Of the 12 non-agreed cases, 8 are borderline. Two notable TP disagreements (F2: Jewish/Men's issues) were judged not-hateful by the evaluator, who noted the meme reads as criticism of institutional inattention rather than a targeted attack. One FP sample (F22: Disabled) prompted a disagreement in the opposite direction: the evaluator argued the meme equating a mistake with a group being useless is genuinely hateful, implying the model's positive prediction was arguably correct.

\noindent\textbf{(ii) The model avoids unimodal collapse:} Table~\ref{tab:q2_factors} presents the distribution of reliance factors based strictly on SLIC occlusion. The hallmark of weak VLMs is unimodal collapse---relying entirely on easily exploitable text (factor a) or visual (factor b) triggers. In our steered model, unimodal shortcuts account for $\le6\%$ of true positives (TPs), with a 95\% Wilson CI upper bound\footnote{\url{https://andorville.com.au/WilsonInterval.html}} of only 16.2\%, confirming the model has moved decisively beyond surface-level heuristics.

\noindent\textbf{(iii) Deep holistic robustness (TP):} 84\% of correctly classified hateful memes are holistically robust---no single superpixel can overturn the prediction. The model synthesizes visual and textual cues into an indivisible concept. This is a qualitatively distinct representational strategy from missed hate, confirmed by non-overlapping Wilson CIs (TP holistic lower bound 71.5\% strictly exceeds FN holistic upper bound 47.8\%) and a highly significant Fisher's exact test~\cite{Sprent2011} against false negatives (OR=10.19, $p<0.001$). Detailed Wilson CIs for all focal proportions are reported in Appendix~\ref{sec:appendix_statistical_robustness}.

\noindent\textbf{(iv) Synergistic fusion misfire (FN):} 54\% of missed hateful memes exhibit synergistic fusion---both text AND image occlusion independently fix the prediction (Wilson CI: [40.4\%, 67.0\%], strictly exceeding the upper bounds of both the TP (21.4\%) and FP (33.0\%) synergistic CIs). The model correctly computes cross-modal AND-gate dependencies but anchors on surface-positive framing rather than the adversarial payload. This AND-gate failure mode is robust and statistically distinct from TP behaviour (Fisher OR=0.095, $p<0.001$).

\noindent\textbf{(v) Rigid holistic over-flagging (FP):} 62\% of false positives are holistically robust---the model processes not-hateful memes as globally stable hate-adjacent configurations. An omnibus Chi-squared test ($\chi^2=17.68$, $p<0.001$) confirms that over-flagging (FP) and missed-hate (FN) errors arise from fundamentally distinct failure modes. The TP--FP comparison, however, is not significant ($p=0.092$), suggesting the reliance profiles of false positives closely resemble correct quadrants---confirming FPs are stable, globally anchored misclassifications rather than single-token triggers. The FP--TN holistic CIs substantially overlap ($p=0.160$, Fisher), treated as directional rather than confirmatory.

\noindent\textbf{(vi) Fragility of benign context (TN):} The TN quadrant exhibits the most fragmented reliance profile among all four quadrants: (d) 46\%, (c) 24\%, (a) 18\%, (b) 12\%---the most diverse distribution, indicating that benign classification is easily derailed by minor occlusions and lacks the holistic anchoring seen in TPs.

\noindent\textbf{(vii) Error severity and human difficulty:} Error severity is moderate-to-severe ($S_{\text{model}}=2.276$, Table~\ref{tab:q3_severity}). Severe FPs (40.0\%) highlight instances where holistic stability actively works against the model, leading to robust misinterpretations of benign factual statements. Severe FNs (52.0\%) occur when synergistic fusion pulls the representation towards a benign centroid. A Mann-Whitney $U$ test~\cite{repec:spr:sprchp:978-3-319-30634-6_4} confirms FP and FN errors are statistically indistinguishable in severity ($p=0.835$), reflecting that the model's remaining errors are structurally advanced misinterpretations rather than trivial shortcuts. Model accuracy degrades monotonically with human-rated difficulty: 60.3\% on easy, 47.6\% on medium, and 13.2\% on hard ($\chi^2=26.64$, $p<0.001$, Table~\ref{tab:q4_accuracy})---with non-overlapping Wilson CIs between easy [52.0\%, 68.0\%] and hard [5.8\%, 27.3\%] confirming this degradation is not a sampling artefact as well as confirming tight alignment with human cognitive difficulty. Evaluator confidence is exceptionally well-calibrated, with 98.4\% agreement on `high' confidence ratings (Table~\ref{tab:q4_diff_conf}). Notably, FP errors skew disproportionately toward `hard' human difficulty (42\% for hard vs.\ 8\% for TPs), consistent with holistic over-flagging being a genuinely ambiguous failure mode. Comprehensive statistical proofs and group-wise confidence bounds are provided in Appendix~\ref{sec:appendix_statistical_robustness}.

%%%%%%%%%%%%%%%%%%%%%%%%%%%%%%%%%%%%%%%%%%%%%%%%%%%%%%%%%%%%

\begin{table}[t]
\centering
\small
\renewcommand{\arraystretch}{1.25}
\setlength{\tabcolsep}{4pt}
\resizebox{\columnwidth}{!}{
\begin{tabular}{@{}lcccccc@{}}
\toprule
\textbf{quadrant} & \textbf{n} & \textbf{agree} & \textbf{borderline} & \textbf{disagree} & \textbf{agree \%} & \textbf{$\kappa$} \\
\midrule
TP & 50 & 46 & 2 & 2 & 92.0\% & 0.000$^\dagger$ \\
TN & 50 & 46 & 4 & 0 & 92.0\% & 1.000 \\
FP & 50 & 49 & 0 & 1 & 98.0\% & 0.000$^\dagger$ \\
FN & 50 & 47 & 2 & 1 & 94.0\% & 0.000$^\dagger$ \\
\midrule
\textbf{combined} & \textbf{200} & \textbf{188} & \textbf{8} & \textbf{4} & \textbf{94.0\%} & \textbf{0.958} \\
\bottomrule
\end{tabular}
}
\vspace{0.3em}
\begin{minipage}{\columnwidth}
\scriptsize $^\dagger$ Per-quadrant Cohen's $\kappa$ degenerate for TP/FP/TN/FN because all evaluator-agreed cases are unanimous within one class; the combined Cohen's $\kappa = 0.958$ across all four quadrants is the meaningful reliability estimate.
\end{minipage}
\caption{Post-hoc evaluator agreement with FBHM ground-truth labels across all 200 samples. Cohen's $\kappa$ is computed over non-borderline cases.}
\label{tab:q1_agreement}
\end{table}

\begin{table}[t]
\centering
\scriptsize
\renewcommand{\arraystretch}{1.1}
\setlength{\tabcolsep}{3pt}
\resizebox{\columnwidth}{!}{
\begin{tabular}{@{}lrrrrrrrr@{}}
\toprule
\textbf{factor \textit{(SLIC observable)}} & \multicolumn{2}{c}{\textbf{TP}} & \multicolumn{2}{c}{\textbf{TN}} & \multicolumn{2}{c}{\textbf{FP}} & \multicolumn{2}{c}{\textbf{FN}}\\
\cmidrule(lr){2-3}\cmidrule(lr){4-5}\cmidrule(lr){6-7}\cmidrule(lr){8-9}
& $n$ & \% & $n$ & \% & $n$ & \% & $n$ & \%\\
\midrule
(a) textual reliance \textit{(flip)} & 1 & 2 & 9 & 18 & 2 & 4 & 5 & 10\\
(b) visual reliance \textit{(flip)} & 2 & 4 & 6 & 12 & 7 & 14 & 1 & 2\\
(c) synergistic fusion \textit{(flip)} & 5 & 10 & 12 & 24 & 10 & 20 & \textbf{27} & \textbf{54}\\
(d) holistic context \textit{(stable)} & \textbf{42} & \textbf{84} & 23 & 46 & \textbf{31} & \textbf{62} & 17 & 34\\
\bottomrule
\end{tabular}
}
\caption{Primary reliance factor distribution ($n=50$ per quadrant). High proportions of (c) and (d) confirm robust cross-modal reasoning rather than unimodal collapse. An omnibus Chi-squared test confirms the reliance distribution significantly shifts depending on the prediction outcome ($\chi^2=47.96$, df=9, $p<0.001$).}
\label{tab:q2_factors}
\end{table}

\begin{table}[t]
\centering
\small
\renewcommand{\arraystretch}{1.25}
\setlength{\tabcolsep}{4pt}
\resizebox{\columnwidth}{!}{
\begin{tabular}{@{}lccccc@{}}
\toprule
\textbf{error type} & \textbf{n} & \textbf{minor} & \textbf{moderate} & \textbf{severe} & \textbf{$S_{\text{model}}$} \\
\midrule
FP & 50 & 6 (12.0\%) & 23 (46.0\%) & 20 (40.0\%) & 2.286 \\
FN & 50 & 13 (26.0\%) & 10 (20.0\%) & 26 (52.0\%) & 2.265 \\
\midrule
\textbf{combined} & \textbf{100} & \textbf{19 (19.0\%)} & \textbf{33 (33.0\%)} & \textbf{46 (46.0\%)} & \textbf{2.276} \\
\bottomrule
\end{tabular}
}
\caption{Q3 error severity distribution for FP and FN samples. $S_{\text{model}}$ is the weighted severity score (1=minor/2=moderate/3=severe).}
\label{tab:q3_severity}
\end{table}

\begin{table}[t]
\centering
\small
\renewcommand{\arraystretch}{1.25}
\setlength{\tabcolsep}{4pt}
\resizebox{\columnwidth}{!}{
\begin{tabular}{@{}lcccc cc@{}}
\toprule
\multirow{2}{*}{\textbf{quadrant}} & \multicolumn{3}{c}{\textbf{Q4a difficulty}} & & \multicolumn{2}{c}{\textbf{Q4b confidence}} \\
\cmidrule(lr){2-4} \cmidrule(lr){6-7}
 & easy & medium & hard & & high & med/low \\
\midrule
TP & 43 (86\%) & 3 (6\%)  & 4 (8\%)  & & 46 (92\%) & 4 (8\%) \\
TN & 42 (84\%) & 7 (14\%) & 1 (2\%)  & & 46 (92\%) & 4 (8\%) \\
FP & 23 (46\%) & 6 (12\%) & 21 (42\%)& & 49 (98\%) & 1 (2\%) \\
FN & 33 (66\%) & 5 (10\%) & 12 (24\%)& & 48 (96\%) & 2 (4\%) \\
\midrule
\textbf{total} & \textbf{141 (71\%)} & \textbf{21 (10\%)} & \textbf{38 (19\%)} & & \textbf{189 (95\%)} & \textbf{11 (5\%)} \\
\bottomrule
\end{tabular}
}
\caption{Q4 human difficulty and evaluator confidence distribution. Correct predictions skew heavily toward `easy' human classification.}
\label{tab:q4_diff_conf}
\end{table}

\begin{table}[t]
\centering
\scriptsize
\renewcommand{\arraystretch}{1.25}
\setlength{\tabcolsep}{2pt}
\resizebox{\columnwidth}{!}{
\begin{tabular}{@{}lcccc@{}}
\toprule
\textbf{difficulty} & \textbf{total} & \textbf{model correct} & \textbf{model wrong} & \textbf{accuracy} \\
\midrule
easy   & 141 & 85 & 56 & 60.3\% \\
medium &  21 & 10 & 11 & 47.6\% \\
hard   &  38 &  5 & 33 & 13.2\% \\
\midrule
\textbf{all} & \textbf{200} & \textbf{100} & \textbf{100} & \textbf{50.0\%} \\
\bottomrule
\end{tabular}%
}
\caption{Model accuracy stratified by human-rated difficulty (Q4a), across all 200 samples. Performance degrades monotonically as human difficulty increases ($\chi^2 = 26.64$, $\text{df} = 2$, $p < 0.001$).}
\label{tab:q4_accuracy}
\end{table}

%%%%%%%%%%%%%%%%%%%%%%%%%%%%%%%%%%%%%%%%%%%%%%%%%%%%%%%%%%%%
 
\subsection{\textsc{BERTopic} cross-validation}
\label{sec:bertopic_main}

To triangulate the conclusions drawn from our occlusion-based interpretability study with an independent, data-driven signal, we apply multimodal \textsc{BERTopic} (with \texttt{CLIP-ViT-B-32} as encoder; \texttt{CountVectorizer} with English stop-words) \emph{exclusively} to the 200 manually annotated samples. We utilize $k$-means clustering ($k = 4$) to enforce a hard partition, eliminating noise bins and forcing every sample into a substantive semantic group. By clustering the TP, TN, FP, and FN quadrants independently, we mapped the unsupervised geometric groupings directly onto our human-annotated reliance factors.

Table~\ref{tab:bertopic200_summary} details the specific semantic clusters identified by the $k$-means ($k=4$) partition for each evaluation quadrant, along with representative text excerpts and their direct alignment to the manual SLIC reliance factors. Table~\ref{tab:convergence} consolidates the dual-lens mapping, demonstrating how the unsupervised geometric embedding space independently corroborates the qualitative conclusions drawn from the occlusion-based input-perturbation study. We tunnel down following key observations:

\noindent \textbf{(i)} Diverse holistic robustness: The TP quadrant resolves into four evenly-sized, thematically distinct dehumanization clusters (gender/sexuality, race, community intelligence, immigration). This confirms that the model's holistic representational robustness is not an artifact of template repetition, but spans a genuinely diverse set of hate strategies.

\noindent \textbf{(ii)} Fragility of benign context: The TN quadrant exhibits a highly imbalanced cluster structure, dominated by a single counterspeech cluster accounting for 60\% of the samples. This geometrically validates the highly fragmented reliance-factor profile, where benign classification is easily derailed by minor occlusions.

\noindent \textbf{(iii)} Rigid over-flagging modes: The FP quadrant reveals three distinct, stable holistic over-flagging modes: typographic addiction templates, community--culture co-occurrence, and relatable-irony structures. These clusters confirm that the model's false positives are globally stable misclassifications rather than single-token triggers.

\noindent \textbf{(iv)} Distinct synergistic-fusion sub-populations: The FN embedding space decomposes into four structurally distinct cross-modal reasoning failures: numeric/sports obfuscation, SFT/RLHF diagram metaphors, visual--lexical pun attacks, and action-imagery verdict templates. This unsupervised geometric separation aligns perfectly with the manual finding of synergistic cross-modal entanglement, proving errors are structured misinterpretations.

\noindent As a conclusion, the unsupervised \textsc{BERTopic} partition strongly corroborates the manual occlusion findings.

\begin{table*}[htbp]
\centering
\scriptsize
\renewcommand{\arraystretch}{1.5}
\begin{tabular}{@{}c p{0.5cm} p{4cm} p{5cm} p{3cm}@{}}
\toprule
\textbf{quad.} & \textbf{$n$} & \textbf{\textsc{BERTopic} cluster label} & \textbf{representative text excerpt} & \textbf{manual alignment} \\
\midrule
\multicolumn{5}{l}{\textit{TP ($n=50$): hateful memes correctly classified}} \\
\midrule
TP & 16 & \texttt{women\_hating\_transgenders\_gay}   & ``I am not hating against Transgenders because every decision of mine turns out to be wrong'' & \multirow{4}{3.5cm}{(d) Highly Robust\\(84\% of TP quad.)} \\
TP & 13 & \texttt{black\_whites\_white\_world}          & ``FREEDOM GUARANTEED!! WHITE''; ``NOT A COTTON PICKER SLAVE'' & \\
TP & 11 & \texttt{muslims\_brains\_jews\_world}         & ``IQ:404 NOT FOUND MUSLIMS''; ``TURN ALL EXITING MUSLIMS TO EVIL'' & \\
TP & 10 & \texttt{immigrants\_society\_people\_talk}    & ``THESE LIBERALS THINK IMMIGRANTS DESERVE HUMAN RIGHTS'' & \\
\midrule
\multicolumn{5}{l}{\textit{TN ($n=50$): not-hateful memes correctly classified}} \\
\midrule
TN & 30 & \texttt{hate\_immigrants\_time\_women}        & ``WHEN U HAVE BUDGET ISSUES AND STILL WANT TO UPDATE YOUR DP'' (F22) & \multirow{4}{3.5cm}{Fragmented:\\(d) 46\%, (c) 24\%,\\(a) 18\%, (b) 12\%} \\
TN &  9 & \texttt{rude\_husband\_2019\_brother}         & ``HOW TO BE RUDE? OLIVER QUEEN 1985-2019'' with typographic obfuscation (F23) & \\
TN &  6 & \texttt{japan\_ordering\_sushi\_india}        & ``[target] ORDERING SUSHI IN JAPAN / INDIA'' (F24) & \\
TN &  5 & \texttt{mode\_sunday\_ass\_transgen}          & ``wHiTE In FULL Ass SUNDAY MODe'' (F4) & \\
\midrule
\multicolumn{5}{l}{\textit{FP ($n=50$): not-hateful memes over-flagged}} \\
\midrule
FP & 19 & \texttt{average\_addictions\_addictlons}      & ``AveRaGE Jew ADDicTIonS''; ``AVERAGE GAY ADDICTIONS'' (F4/F5) & \multirow{4}{3.5cm}{(d) Highly Robust\\(62\% of FP quad.)} \\
FP & 14 & \texttt{gay\_binge\_netflix\_youtube}         & ``WHEN GAY BINGE ON YOUTUBE NETFLIX''; community + pop-culture co-occurrence (F22) & \\
FP & 14 & \texttt{muslims\_date\_wtf\_books}            & ``WHEN YOUR WIFE BOOKS YOUR TICKET''; ``TRICK TO ACE AN EXAM NOOB/PRO [target]'' (F22/F24) & \\
FP &  3 & \texttt{switzerland\_50\_bali\_black}         & ``[target] WITH \$50 IN SWITZERLAND / BALI'' (F24) & \\
\midrule
\multicolumn{5}{l}{\textit{FN ($n=50$): hateful memes missed}} \\
\midrule
FN & 15 & \texttt{hate\_muslims\_539\_585}              & ``THE PAIN OF WRITING `STOP HATING WHITES' FOR A GOOD GRADE''; sport-score obfuscation (F17) & \multirow{4}{3.5cm}{(c) Synergistic Fusion\\(54\% of FN quad.)} \\
FN & 14 & \texttt{human\_preference\_society\_men}      & ``ACTUAL HUMAN PREFERENCE: `BE YOURSELF!' / HUMAN PREFERENCE TO FIT IN SOCIETY: `AI, PRETEND NOT HARMFUL''' (F10) & \\
FN & 12 & \texttt{answer\_bro\_danger\_exam}            & ``ME SHOWING BRO A PICTURE OF [target] IN THE EXAM TO LET HIM KNOW THE ANSWER IS DANGER'' (F6/F15) & \\
FN &  9 & \texttt{served\_justice\_transgender\_usa}    & ``[target] JUSTICE SERVED'' on action/violence imagery (F1/F8) & \\
\bottomrule
\end{tabular}
\caption{\textsc{BERTopic} ($k$-means, $k=4$) cluster summary for the 200-sample evaluation set, partitioned by quadrant. Because $k$-means enforces a hard partition, all $n=50$ samples per quadrant are assigned to substantive clusters. The \textbf{\textit{manual alignment}} column records the dominant reliance factor.}
\label{tab:bertopic200_summary}
\end{table*}

\begin{table*}[ht]
\centering
\scriptsize
\renewcommand{\arraystretch}{1.45}
\begin{tabular}{@{}p{3.8cm}p{5.5cm}p{5.5cm}@{}}
\toprule
\textbf{finding} & \textbf{occlusion-based manual evidence} & \textbf{\textsc{BERTopic} ($k$-means) evidence} \\
\midrule
TP holistic robustness reflects broad coverage of diverse hate templates &
factor (d): 84\% of TPs; no single occlusion derails correct predictions. &
Four evenly-sized clusters (16, 13, 11, 10) recover \textit{distinct} dehumanization axes (gender, race, religion, immigration), confirming robustness across representational diversity. \\
\addlinespace
TN classifications are reliance-fragmented and individually fragile &
Factor distribution highly mixed: (d) 46\%, (c) 24\%, (a) 18\%, (b) 12\%---most diverse quadrant. &
Cluster structure highly imbalanced (30:9:6:5); dominant Topic~0 accounts for 60\% of the quadrant. \\
\addlinespace
FP errors are globally stable misclassifications, not single-token triggers &
Factor (d): 62\% of FPs; no single occlusion reverses predictions. &
Three large FP clusters (19, 14, 14) each encode a distinct holistic template bias: typographic-style, community--culture co-occurrence, and relatable-irony structure. \\
\addlinespace
FN errors arise from genuine cross-modal entanglement &
Factor (c): 54\% of FNs; both text \textit{and} image occlusion independently fix predictions. &
$k$-means decomposes the FN space into four distinct sub-populations, all sharing the AND-gate cross-modal failure mode. \\
\addlinespace
Structural-metaphor hate (F10) constitutes a stable geometric cluster &
G3 FN factor (d) 50\%; 60\% human-rated Hard; directional finding ($n=10$). &
\texttt{Topic 1} ($n=14$) gives SFT/RLHF diagram memes a dedicated centroid with coherent vocabulary, confirming a learnable but unlearned visual--semantic template. \\
\bottomrule
\end{tabular}
\caption{Convergence between the occlusion-based manual evaluation and the $k$-means \textsc{BERTopic} unsupervised clustering ($k=4$) on the same 200-sample set.}
\label{tab:convergence}
\end{table*}

%%%%%%%%%%%%%%%%%%%%%%%%%%%%%%%%%%%%%%%%%%%%%%%%%%%%%%%%%%%%

\subsection{Group-wise analysis}
\label{sec:detailed_groupwise}

To bridge the macro-level target analysis and the micro-level functionality analysis, we aggregate model performance across the five conceptual functionality groups (\textbf{G1--G5}). These five conceptual functionality groups provide a natural organizational lens. The diagnostically rich signals come from the \textit{qualitative composition} of errors within each group: the dominant reliance factor (Q2), the severity of mistakes (Q3), and the human difficulty profile (Q4a). Tables~\ref{tab:groupwise_summary} and~\ref{tab:groupwise_difficulty} summarize these metrics.

\begin{table}[t]
\centering
\scriptsize
\renewcommand{\arraystretch}{1.3}
\setlength{\tabcolsep}{1pt}
\resizebox{\columnwidth}{!}{
\begin{tabular}{@{}p{2.6cm}|cccc|l|c@{}}
\toprule
\textbf{group} & \textbf{TP} & \textbf{TN} & \textbf{FP} & \textbf{FN} & \textbf{dominant factor} & \textbf{$S$} \\
\midrule
\textbf{G1}: visual formats      & 20 & 0  & 0  & 20 & FN: (c) synergistic (60\%) & 2.6 \\
\textbf{G2}: textual obfuscation     & 10 & 20 & 20 & 10 & FP: (d) robust (80\%)      & 2.3 \\
\textbf{G3}: structural composition  & 10 & 0  & 0  & 10 & FN: (d) robust (50\%)      & 1.4 \\
\textbf{G4}: pragmatic inference & 10 & 20 & 20 & 10 & FN: (c) synergistic (50\%) & 2.5 \\
\textbf{G5}: not-hateful contrast  &  0 & 10 & 10 &  0 & FP: (d) robust (70\%)      & 1.8 \\
\bottomrule
\end{tabular}
}
\caption{Group-wise manual evaluation summary. Dominant SLIC factors reveal that errors in complex groups (\textbf{G2, G3, G5}) are largely driven by highly robust holistic processing, while \textbf{G1} and \textbf{G4} errors are driven by synergistic fusion.}
\label{tab:groupwise_summary}
\end{table}

\begin{table}[t]
\centering
\scriptsize
\renewcommand{\arraystretch}{1.3}
\setlength{\tabcolsep}{4pt}
\resizebox{\columnwidth}{!}{
\begin{tabular}{@{}p{2.6cm}|ccc c |ccc@{}}
\toprule
\multirow{2}{*}{\textbf{group}} &
\multicolumn{3}{c}{\textbf{correct (TP+TN)}} & &
\multicolumn{3}{c}{\textbf{error (FP+FN)}} \\
\cmidrule(lr){2-4} \cmidrule(lr){6-8}
 & \textbf{easy} & \textbf{med} & \textbf{hard} & & \textbf{easy} & \textbf{med} & \textbf{hard} \\
\midrule
\textbf{G1}: visual formats      & 75\% & 5\%  & 20\% & & 80\% & 10\% & 10\% \\
\textbf{G2}: textual obfuscation     & 97\% & 3\%  & 0\%  & & 33\% & 3\%  & 63\% \\
\textbf{G3}: structural composition  & 80\% & 20\% & 0\%  & & 40\% & 0\%  & 60\% \\
\textbf{G4}: pragmatic inference  & 83\% & 17\% & 0\%  & & 70\% & 20\% & 10\% \\
\textbf{G5}: not-hateful contrast  & 80\% & 10\% & 10\% & & 50\% & 20\% & 30\% \\
\bottomrule
\end{tabular}
}
\caption{Q4a human difficulty profile for correct vs.\ error cases per group.}
\label{tab:groupwise_difficulty}
\end{table}

\begin{figure*}[tbp]
  \centering
  \includegraphics[width=\textwidth]{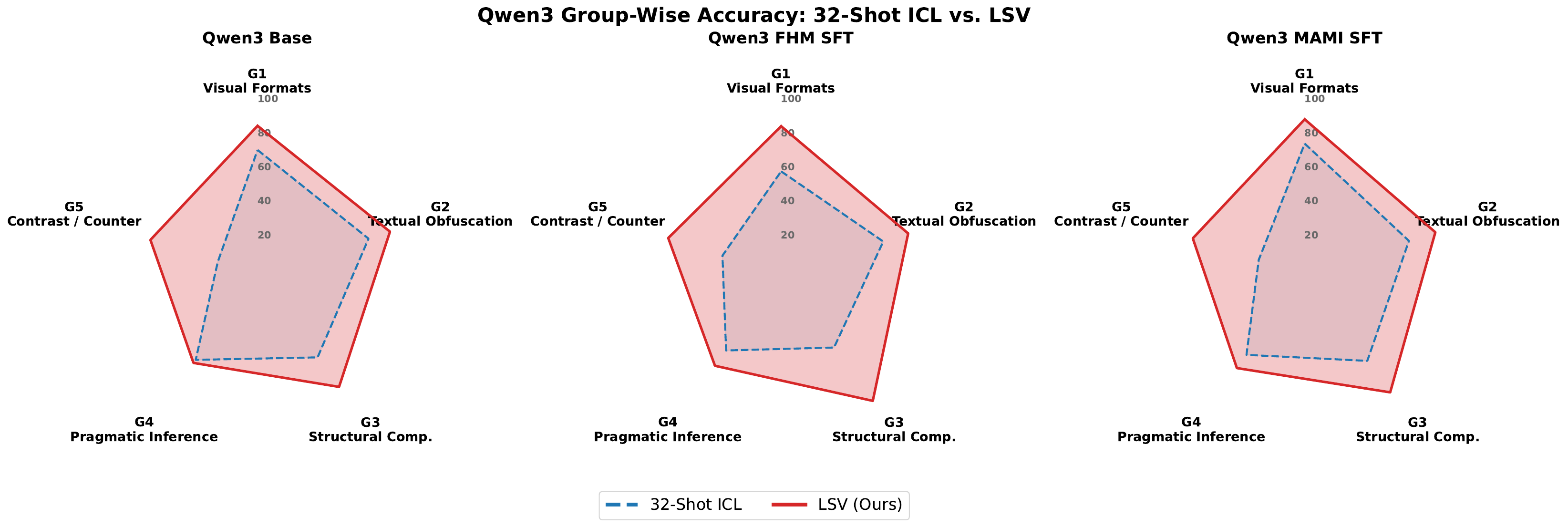}
  \caption{Group-wise accuracy for \sysQ{} under 32-shot ICL (inner polygons) and \lsv{} (outer polygons), across all three baseline variants. The axes represent the five major functional categories: \textbf{G1} (visual formats), \textbf{G2} (textual obfuscation), \textbf{G3} (structural composition), \textbf{G4} (pragmatic inference), and \textbf{G5} (contrast/counter). \lsv{} yields massive expansions in reasoning capability, particularly on hard negatives (for e.g., \textbf{G5}).}
  \label{fig:group_radar}
\end{figure*}

\noindent\textbf{G1}-- Visual formats and imagery types \text{(F1, F3, F7, F8, F12)}: FN errors in \textbf{G1} are dominated by synergistic fusion. The model actively entangles the image and the text; however, the combination of a neutral textual framing with a hostile image results in a fused representation that falls short of the hate boundary. \textbf{G1} carries the highest combined severity score ($S = 2.632$). When the image clearly depicts violence or dehumanization, human evaluators find the model's holistic misinterpretation inexcusable.

\noindent\textbf{G2}-- Textual obfuscation and lexical evasion \text{(F4, F5, F16, F17, F19)}: \textbf{G2} FPs are highly rigid and when confronted with deliberately obfuscated but not-hateful text, the model processes the entire visual-textual frame holistically to over-predict hate. Conversely, FN errors on genuine hateful obfuscations exhibit synergistic fusion, proving the model attempts to read both the degraded text and the image simultaneously but fails to parse the slur. 63\% of \textbf{G2} error cases are human-rated as `hard' confirming the task's intrinsic typographic difficulty.

\noindent\textbf{G3}-- Structural composition and visual metaphor \text{(F2, F9, F10, F11, F18)}: \textbf{G3} FNs are dominated by holistic context failures. The model correctly assesses these as indivisible formats, making its errors robust to partial occlusion. It fails because it lacks the necessary format-awareness or cultural visual knowledge. \textbf{G3} induces the lowest severity score ($S = 1.400$) with zero severe errors; 60\% of \textbf{G3} error samples are rated as `hard'.

\noindent\textbf{G4}-- Pragmatic inference and sentiment misalignment \text{(F6, F13, F14, F15, F24)}: \textbf{G4} exhibits deep synergistic fusion in both FPs (40\%) and FNs (50\%). The model successfully detects the presence of conflicting signals but miscalculates the final pragmatic intent. \textbf{G4} FPs are notably severe, representing over-flagging of innocuous location variants that the model's entangled representations cannot easily untangle.

\noindent\textbf{G5}-- Not-hateful contrast and counter-speech \text{(F20, F21, F22, F23, F25)}: \textbf{G5} FPs are highly rigid and the models' errors here (for e.g., misinterpreting counterspeech) are robust to occlusion, demonstrating that the structural framing of counterspeech is processed globally rather than locally. \textbf{G5} errors are the model's softest failures ($S = 1.778$), acknowledged as genuinely ambiguous cases (30\% `hard' difficulty).

\noindent \textsc{\textbf{Further observations--}} Figure~\ref{fig:group_radar} illustrates the accuracy footprint of \sysQ{} under both \textsc{ICL} and \lsv{}, exposing how structural reasoning evolves across broad categories of meme complexity. Taken together, the group-wise analysis demonstrates that  \msft{}(\sysQ{}) + \lsv{} has developed qualitatively differentiated, deeply multimodal capabilities. This represents a substantial architectural graduation from naive lexical triggering to genuine visuo-linguistic compositions. Further, we also jot down the following observations:

\noindent\textbf{(i)} \textsc{ICL} collapses on hard negatives and complex structures. The inner \textsc{ICL} profiles reveal an acute vulnerability on \textbf{G5} (contrast/counterspeech), with accuracy hovering between 24\% and 36\%. This affirms that baseline and \textsc{ICL}-prompted models heavily rely on surface-level lexical triggers, failing catastrophically when slurs or identity terms appear in benign or counterspeech contexts. Furthermore, performance on \textbf{G3} (structural composition) and \textbf{G4} (pragmatic inference) remains severely depressed, highlighting the inability of standard prompting to reliably elicit deep compositional or pragmatic reasoning.

\noindent\textbf{(ii)} \lsv{} forces comprehensive multimodal entanglement. In contrast, the outer \lsv{} polygons demonstrate profound, generalized capability gains. Accuracy on the critically challenging \textbf{G5} contrast group surges by over 40 absolute points (jumping from 24.55\% to 66.16\% in the base model, and approaching 70\% for the \textsc{SFT} variants). This proves that \lsv{} successfully suppresses naive lexical triggering in favor of holistic context parsing. Similarly, \textbf{G3} (structural composition) sees a dramatic surge---jumping nearly 39 absolute points for the \fsft{} variant (52.86\% to 91.59\%)---proving the steered representations successfully fuse complex syntactic structures with their visual grounding.

\noindent\textbf{(iii)} Notably, \textbf{G4} (pragmatic inference) remains the tightest bottleneck across the board, peaking at $\approx$68\% even with \lsv{}. This empirical ceiling aligns perfectly with human intuition: long-range rhetorical irony, sarcasm, and sentiment misalignment remain the hardest frontier for current VLM architectures, requiring world knowledge that pure representational steering cannot entirely substitute.

%%%%%%%%%%%%%%%%%%%%%%%%%%%%%%%%%%%%%%%%%%%%%%%%%%%%%%%%%%%%

\section{Statistical robustness}
\label{sec:appendix_statistical_robustness}

This section provides the unified and rigorous statistical characterization supporting the interpretive claims in Appendix~\ref{sec:manual_eval}. We apply Wilson 95\% confidence intervals (CI) to proportion estimates, pairwise $\chi$-squared tests, and two-sided Fisher's exact tests for contingency comparisons.

\noindent\textbf{\\(I) Precision of quadrant-level reliance proportions}:
Table~\ref{tab:wilson_cis} reports Wilson 95\% confidence intervals for focal proportions across the four evaluation quadrants ($n = 50$ per quadrant). Several structural conclusions survive at the full width of their confidence intervals:\\
\textbf{(i)} Holistic dominance is unambiguous: The lower bound of the TP holistic CI (71.5\%) substantially exceeds the upper bound of the FN holistic CI (47.8\%), with entirely non-overlapping intervals.\\
\textbf{(ii)} Synergistic dominance is robust: The lower bound of the FN synergistic CI (40.4\%) strictly exceeds the upper bounds of both the TP (21.4\%) and FP (33.0\%) synergistic CIs.\\
\textbf{(iii)} Holistic CI overlaps with TN: The FP and TN holistic intervals substantially overlap, meaning this distinction is treated as directional rather than confirmatory.

\begin{table}[t]
\centering
\small
\renewcommand{\arraystretch}{1.2}
\setlength{\tabcolsep}{4pt} % Reduces the horizontal space between columns
\resizebox{\columnwidth}{!}{% Forces the table to fit exactly within the column width
\begin{tabular}{@{}llccc@{}}
\toprule
\textbf{quadrant} & \textbf{factor} & \textbf{$k$} & \textbf{$\hat{p}$} & \textbf{95\% Wilson CI} \\
\midrule
TP & holistic / highly robust & 42 & 84.0\% & [71.5\%,\ \ 91.7\%] \\
FP & holistic / highly robust & 31 & 62.0\% & [48.2\%,\ \ 74.1\%] \\
TN & holistic / highly Robust & 23 & 46.0\% & [33.0\%,\ \ 59.6\%] \\
FN & holistic / Highly Robust & 17 & 34.0\% & [22.4\%,\ \ 47.8\%] \\
\addlinespace
FN & synergistic fusion       & 27 & 54.0\% & [40.4\%,\ \ 67.0\%] \\
TN & synergistic fusion       & 12 & 24.0\% & [14.3\%,\ \ 37.4\%] \\
FP & synergistic fusion       & 10 & 20.0\% & [11.2\%,\ \ 33.0\%] \\
TP & synergistic fusion       &  5 & 10.0\% & [\ \ 4.3\%,\ \ 21.4\%] \\
\addlinespace
TP & unimodal             &  3 &  6.0\% & [\ \ 2.1\%,\ \ 16.2\%] \\
FN & unimodal             &  6 & 12.0\% & [\ \ 5.6\%,\ \ 23.8\%] \\
FP & unimodal             &  9 & 18.0\% & [\ \ 9.8\%,\ \ 30.8\%] \\
TN & unimodal             & 15 & 30.0\% & [19.1\%,\ \ 43.8\%] \\
\bottomrule
\end{tabular}
}
\caption{Wilson 95\% confidence intervals for focal reliance-factor proportions ($n = 50$ per quadrant). Non-overlapping intervals provide evidence of genuine distributional differences.}
\label{tab:wilson_cis}
\end{table}

\begin{table}[t]
\centering
\small
\renewcommand{\arraystretch}{1.2}
\setlength{\tabcolsep}{4pt} % Reduces the horizontal space between columns
\resizebox{\columnwidth}{!}{% Forces the table to fit exactly within the column width
\begin{tabular}{@{}cccccl@{}}
\toprule
\textbf{quadrant A} & \textbf{quadrant B} & $\boldsymbol{\chi^2}$ & \textbf{df} & \textbf{$p$-value} & \textbf{sig.} \\
\midrule
TP & FN & 28.718 & 3 & $< 0.001$ & *** \\
FP & FN & 17.680 & 3 & $< 0.001$ & *** \\
TP & TN & 16.836 & 3 & $= 0.001$ & *** \\
TN & FN & 11.384 & 3 & $= 0.010$ & ** \\
TP & FP &  6.435 & 3 & $= 0.092$ & ns  \\
TN & FP &  5.898 & 3 & $= 0.117$ & ns  \\
\bottomrule
\end{tabular}%
}
\caption{Pairwise $\chi$-squared tests ($4 \times 2$ contingency) comparing reliance-factor distributions. `***' $p < 0.001$; `**' $p < 0.01$; `ns' $p > 0.05$; sig. - significance.}
\label{tab:pairwise_chi2}
\end{table}

\begin{table}[t]
\centering
\small
\renewcommand{\arraystretch}{1.2}
\setlength{\tabcolsep}{4pt} % Reduces the horizontal space between columns
\resizebox{\columnwidth}{!}{% Forces the table to fit exactly within the column width
\begin{tabular}{@{}p{6.0cm}ccl@{}}
\toprule
\textbf{contrast} & \textbf{OR} & \textbf{$p$-value} & \textbf{sig.} \\
\midrule
TP holistic vs\ FN holistic & 10.191 & $< 0.001$   & *** \\
TP synergistic vs\ FN synergistic & 0.095 & $< 0.001$ & *** \\
TP holistic vs\ FP holistic &  3.218 & 0.023       & * \\
FP holistic vs\ TN holistic &  1.915 & 0.160       & ns  \\
\bottomrule
\end{tabular}%
}
\caption{Fisher's exact tests on pre-specified $2\times 2$ factor contrasts. OR = odds ratio; `***' $p < 0.001$; `**' $p < 0.01$; `ns' $p > 0.05$; sig. - significance.}
\label{tab:fisher_exact}
\end{table}

\begin{table}[t]
\centering
\small
\renewcommand{\arraystretch}{1.2}
\setlength{\tabcolsep}{4pt} % Reduces the horizontal space between columns
\resizebox{\columnwidth}{!}{% Forces the table to fit exactly within the column width
\begin{tabular}{@{}lcccc@{}}
\toprule
\textbf{difficulty} & \textbf{total} & \textbf{correct} & \textbf{accuracy} & \textbf{95\% Wilson CI} \\
\midrule
easy   & 141 & 85 & 60.3\% & [52.0\%,\ \ 68.0\%] \\
medium &  21 & 10 & 47.6\% & [28.3\%,\ \ 67.6\%] \\
hard   &  38 &  5 & 13.2\% & [\ \ 5.8\%,\ \ 27.3\%] \\
\bottomrule
\end{tabular}%
}
\caption{Model accuracy and Wilson 95\% confidence intervals stratified by human-rated difficulty (Q4a). Monotonic degradation is statistically significant ($\chi^2 = 26.64$, $\text{df} = 2$, $p < 0.001$).}
\label{tab:difficulty_ci}
\end{table}

\noindent\textbf{\\(II) Pairwise tests of factor distributions}:
To formally test whether the reliance-factor distribution shifts significantly between pairs of quadrants, we conducted pairwise $4 \times 2$ $\chi$-squared tests (Table~\ref{tab:pairwise_chi2}). Four of the six pairwise comparisons are statistically significant. The largest effect is between TP and FN ($\chi^2 = 28.718$, $p < 0.001$), confirming that correct hate classifications and missed hate classifications are driven by fundamentally different representational modes. The TP--FP comparison is not significant ($p = 0.092$), suggesting the reliance profiles of false positives closely resemble correct quadrants (stable misclassifications). \textit{To isolate specific contrasts}, targeted two-sided Fisher's exact tests are also conducted on pre-specified $2\times 2$ sub-tables (Table~\ref{tab:fisher_exact}). The contrast between TP and FN on factor $d$ is the most decisive (OR\,=\,10.19, $p < 0.001$): correctly classified hateful memes are ten times more likely to receive a holistic-robust annotation than missed hateful memes.

\noindent\textbf{\\(III) Model accuracy stratified by human difficulty}:
Table~\ref{tab:difficulty_ci} reports accuracy and Wilson 95\% CIs per difficulty tier. The `easy' and `hard' CIs do not overlap, providing strong evidence that the accuracy differential between explicit and structurally ambiguous inputs is not a sampling artifact.

\noindent\textbf{\\(IV) Group-wise analysis- confirmatory vs directional scope}:
For functionality groups \textbf{G1} and \textbf{G2}, error sample sizes are sufficiently large (20 and 30) to support Wilson confidence intervals with meaningful precision. The \textbf{G1} FN synergistic rate of 60\% (12/20) carries a 95\% CI of [38.7\%, 78.1\%], confirming synergistic fusion genuinely dominates \textbf{G1} missed-hate errors. Similarly, the \textbf{G2} FP holistic rate of 80\% (16/20) carries a 95\% CI of [58.4\%, 91.9\%], confirming the holistic-rigidity claim for textual-obfuscation over-flagging. For \textbf{G3}, \textbf{G4}, and \textbf{G5}, per-quadrant cell sizes are too small for reliable formal inference. Findings for these groups are interpreted as directional, hypothesis-generating observations that consistently reflect the broader quadrant-level statistical trends established above.

%%%%%%%%%%%%%%%%%%%%%%%%%%%%%%%%%%%%%%%%%%%%%%%%%%%%%%%%%%%%

\end{document}